\documentclass[pdflatex,sn-basic]{sn-jnl}

\usepackage{graphicx}%
\usepackage{multirow}%
\usepackage{amsmath,amssymb,amsfonts}%
\usepackage{amsthm}%
\usepackage{mathrsfs}%
\usepackage[title]{appendix}%
\usepackage{xcolor}%
\usepackage{textcomp}%
\usepackage{tipa}%
\usepackage{manyfoot}%
\usepackage{booktabs}%
\usepackage{algorithm}%
\usepackage{algorithmicx}%
\usepackage{algpseudocode}%
\usepackage{listings}%
\usepackage{xcolor}         
\usepackage{soul}
\usepackage{graphicx}       
\usepackage{tikz}           
\usepackage{cancel}
\usepackage[most]{tcolorbox}
\usepackage[capitalise,noabbrev]{cleveref}
\usepackage{lineno}

\theoremstyle{thmstyleone}%
\theoremstyle{thmstyletwo}%

\theoremstyle{thmstylethree}%

\definecolor{VeryLightBlue}{HTML}{E6F7FF}
\definecolor{VeryLightYellow}{HTML}{FFFFE6}
\definecolor{VeryLightGreen}{HTML}{E6FFE6} 

\newtcolorbox{userprompt}{
  breakable,
  colback=red!5!white,
  colframe=purple!60!black,
  width=\dimexpr\linewidth+6cm\relax,
  enlarge left by=-3cm,
  enlarge right by=-3cm,
  boxrule=2pt,
  coltext=black,
  arc=4pt,
  left=1pt,
  right=1pt,
  top=4pt,
  bottom=4pt,
  before skip=8pt,
  after skip=8pt,
}

\newtcolorbox{updatedPrompt}{
  breakable,
  colback=red!5!white,
  colframe=cyan!60!black,
  width=\dimexpr\linewidth+6cm\relax,
  enlarge left by=-3cm,
  enlarge right by=-3cm,
  boxrule=2pt,
  coltext=black,
  arc=4pt,
  left=1pt,
  right=1pt,
  top=4pt,
  bottom=4pt,
  before skip=8pt,
  after skip=8pt,
}

\newtcolorbox{systemprompt}{
  breakable,
  colback=red!3!white,
  colframe=red!90!black,
  width=\dimexpr\linewidth+6cm\relax,
  enlarge left by=-3cm,
  enlarge right by=-3cm,
  coltext=red,
  boxrule=2pt,
  arc=4pt,
  left=6pt,
  right=6pt,
  top=4pt,
  bottom=4pt,
  before skip=8pt,
  after skip=8pt
}

\newtcolorbox{aiprompt}{
  breakable,
  colback=violet!5!white,
  colframe=violet!60!black,
  width=\dimexpr\linewidth+6cm\relax,
  enlarge left by=-3cm,
  enlarge right by=-3cm,
  boxrule=2pt,
  arc=4pt,
  left=6pt,
  right=6pt,
  top=4pt,
  bottom=4pt,
  before skip=8pt,
  after skip=8pt
}

\begin{document}

\title[Virtual Speech Therapist:A Clinician-in-the-Loop AI Agent for Speech Therapy Planning]{\textcolor{black}{Virtual Speech Therapist: A Clinician-in-the-Loop AI Agent for Speech Therapy Planning}}

\author*[1]{\fnm{Shakeel A.} \sur{Sheikh}}\email{shakeelzmail608@gmail.com}

\author*[2]{\fnm{Patrick} \sur{Marmaroli}}\email{patrick.marmaroli@gmail.com, schuller@ieee.org}
\author[3]{\fnm{Md} \sur{Sahidullah}\email{md.sahidullah@tcgcrest.org}}
\author[4]{\fnm{Slim} \sur{Ouni}\email{slim.ouni@loria.fr}}
\author[5]{\fnm{Fabrice} \sur{Hirsch}\email{fabrice.hirsch@univ-montp3.fr}}
\author[6]{\fnm{Gon\c{c}alo} \sur{Leal}\email{goncalo.leal@speechcare.center}}
\author[7]{\fnm{Bj\"orn W.} \sur{Schuller}\email{schuller@ieee.org}
}

\affil[1]{\orgname{DMI, DoPH, Luxembourg Institute of Health},
\orgaddress{\city{Strassen}, \country{Luxembourg}}}

\affil[2]{\orgname{Vocametrix},
\orgaddress{\city{Tallinn}, \country{Estonia}}}

\affil[3]{\orgname{IAI, TCG CREST},
\orgaddress{\state{West Bengal}, \country{India}}}

\affil[4]{\orgname{Universit\'e de Lorraine, CNRS, Inria, LORIA},
\orgaddress{\city{Nancy}, \postcode{54000}, \country{France}}}

\affil[5]{\orgname{Laboratoire Praxiling, Universit\'e Paul-Val\'ery Montpellier 3},
\orgaddress{\country{France}}}

\affil[6]{\orgname{Speechcare  iStutter, Portuguese Catholic University},
\orgaddress{\country{Portugal}}}

\affil[7]{\orgdiv{CHI-Chair of Health Informatics}, \orgname{TUM University Hospital},
\orgaddress{\country{Germany}}}

\affil[7]{\orgdiv{GLAM-Group on Language, Audio, \& Music}, \orgname{Imperial College London},
\orgaddress{\country{United Kingdom}}}


\abstract{\textcolor{black}{This paper presents a proof of concept of the \textit{Virtual Speech Therapist} (VST)\footnote{Code: \href{https://github.com/pmarmaroli/vocametrix-platform/tree/main/python/vstagent}{https://github.com/pmarmaroli/vocametrix-platform/tree/main/python/vstagent}}, an intelligent multi-agent based system designed to support stuttering assessment and individualized therapy planning through artificial intelligence (AI) driven workflows. VST integrates deep learning (DL) based stuttering detection $f_{SD}$ with large language model (LLM) based agentic system to assist clinicians in developing evidence informed therapy plans. The VST system begins with the acquisition and feature extraction of given speech samples, followed by the detection of stuttering types. Building on these speech driven information $\mathcal{S}$, VST initiates an agentic reasoning process in which specialized LLM agents autonomously generate, critique, and iteratively refine individualized therapy plans $\mathcal{P}$. A dedicated \textit{CriticAgent} ($\pi_{ca}$) acts as a structured self-refinement mechanism, reviewing the generated therapy plans $\mathcal{P}$ and identifying areas for potential improvement in consistency, methodological soundness, and alignment with available evidence and professional guidelines. The resulting therapy plan $\mathcal{P}$ is presented then as a draft for clinician review. Clinician feedback can then be incorporated to produce a revised plan $\mathcal{P}$, thereby maintaining a clinician-in-the-loop (CITL) paradigm. An initial evaluation, involving expert speech language pathologist (SLP) examines the clinical interpretability and feasibility of the generated therapy plans. Taken together, the results show the potential of VST to augment clinician-led therapy planning and provide a promising basis for future studies evaluating its integration into real world clinical settings.
An interactive user interface for the proposed prototype is available online at: \href{https://vocametrix.com/ai/stuttering-therapy-planning-agent}{https://vocametrix.com/ai/stuttering-therapy-planning-agent}, facilitating interactive stuttering assessment and stuttering therapy planning.}}


\maketitle

\section{Introduction}

{
\color{black}

Speech is a complex sensorimotor behavior that requires coordinated control of respiratory, phonatory, and articulatory processes across distributed neural networks~\citep{kearney2019articulating}. Disruptions within this intricate system can impair speech fluency, manifesting as speech disorders such as stuttering, which is  described as the involuntary repetitions of sounds, syllables, or monosyllabic words, prolongations, and blocks. These core behaviours may be accompanied by secondary behaviours, such as muscular tension, eye blinking, head movements or avoidance~\citep{guitar2018stuttering}. In this paper, stuttering refers to developmental stuttering, classified in ICD-11 as a developmental speech fluency disorder (6A01.1): a persistent disruption of the normal flow and rate of speech, with onset in the developmental period, shown as repetitions and prolongations of sounds, syllables, words and phrases, blocks, and word avoidance or substitutions, and causing functional impairment~\citep{who2025icd11}. The experience of stuttering may also extend beyond observable speech behaviors, encompassing cognitive and emotional experiences such as anticipation, anxiety, frustration, and negative reactions to communication, which may in turn influence social participation and quality of life~\citep{beilby2014psychosocial, tichenor2019stuttering}.

Stuttering typically emerges during the preschool years, most commonly between two and four years of age. Its onset during early childhood is often followed by spontaneous recovery, as demonstrated by a prospective community cohort in which the cumulative incidence of stuttering reached 11.2\% by age four~\citep{reilly2020natural}, while follow-up of the same cohort found that 65\% of children had recovered by age seven~\citep{kefalianos2017history}. Despite this substantial rate of recovery during childhood, stuttering persists into adulthood for some individuals, with adult prevalence estimates ranging from approximately 0.6\% to 1.0\%, depending on the identification methods used and whether covert stuttering is taken into account~\citep{gattie2025adulta, gattie2025adultb}.

Stuttering has a multifactorial neurodevelopmental basis and has been associated with atypicalities in neural circuits involved in speech sensorimotor integration and movement initiation~\citep{chang2019functional, etchell2018systematic}. Its development and persistence are thought to reflect interactions among neural, motor, linguistic, and emotional factors~\citep{smith2017stuttering}. This heterogeneity underscores the importance of individualized assessment and intervention planning~\citep{sheikhbahaei2023stuttering}. In this context, computational approaches that can organize speech derived information may offer a means of supporting clinicians in developing individualized therapy plans, provided that their outputs remain subject to clinical interpretation and oversight~\citep{tayebi2026privacy}.

Clinical assessment of stuttering involves identifying and quantifying different types of stuttering events within speech samples. However, manual segmentation and annotation of these events can be time consuming, labor intensive, and resource intensive. Recent advances in automated speech analysis have shown potential for assisting in the identification and characterization of stuttering events~\citep{bayerl2023classification, sheikh2022mlstuttering}. By providing structured information about observed speech characteristics, such automated analyses may complement clinical assessment and contribute to the information considered when developing intervention therapy plans~\citep{CHEN201698}. Nevertheless, automated identification of stuttering events alone does not determine which intervention strategies are appropriate for an individual, as therapy planning requires the integration of speech related findings with clinical expertise and patient specific considerations~\citep{yaruss2021consensus}.

To explore how such automated speech information could be incorporated into intervention planning, we propose a \emph{Virtual Speech Therapist} (VST), a multi-agentic LLM based framework that uses automated stuttering detection $f_{SD}$ results as one source of information for generating candidate therapy plans targeting speech motor aspects of intervention under clinician supervision. The generated plans are intended to function as decision support outputs rather than substitutes for clinical judgment. Their \textit{interpretation} and \textit{use} require consideration of peer-reviewed evidence, the individual’s values and goals, and the broader clinical and practice context. Importantly, speech motor intervention represents only one component of stuttering care; individualized intervention may also need to consider cognitive and emotional experiences, communication related participation, and other factors relevant to the individual's goals and needs. Taken together, these findings provide preliminary evidence that VST may augment clinician-led therapy planning by integrating automated speech related information into the generation of candidate intervention plans. They provide a basis for further investigation of how such AI-assisted systems might be incorporated into real world clinical workflows while remaining subject to appropriate clinician oversight and decision making.

}

{\color{black}
\section{Automated Stuttering Detection}
Stuttering assessment has historically relied on manual auditory perceptual judgments conducted by trained \emph{speech language pathologists} (SLPs)~\citep{riley1972stuttering}. Although such assessments remain clinically valuable, they are time consuming and costly, and their reliance on human perception can introduce inter and intra rater variability~\citep{guitar2018stuttering, yairi2013school, lea:2021, guitar2024stuttering}. These limitations have motivated the development of computational approaches that can assist clinicians by providing more objective and scalable analyses of speech. Early efforts toward automated stuttering detection primarily adopted conventional machine learning pipelines based on handcrafted acoustic features~\citep{sheikh2022mlstuttering}. In these approaches, models such as support vector machines and hidden Markov models were trained on features including jitter, shimmer, and spectral characteristics intended to capture the acoustic and prosodic disruptions associated with stuttered speech~\citep{sheikh2022mlstuttering}. While these studies established the feasibility of computational stuttering detection, their effectiveness was inherently dependent on the quality and completeness of manually engineered representations, limiting their ability to capture the complex variability of stuttered speech across speakers and recording conditions.

The emergence of deep learning (DL) subsequently shifted this paradigm by enabling models to learn task relevant representations directly from raw speech signals or minimally processed acoustic representations. Convolutional neural networks, for example, have been applied to time-frequency representations such as spectrograms, allowing DL models to learn local spectral-temporal patterns associated with different types of stuttering~\citep{kourkounakis2020stuttering, kourkounakis2021fluentnet,sheikh2022robust,s2023advancing, sheikh2021stutternet}. Because stuttering unfolds dynamically over time, researchers have also explored recurrent architectures, particularly bidirectional long short-term memory networks, to capture temporal dependencies and contextual information extending beyond individual acoustic frames~\citep{kourkounakis2021fluentnet}. As research progressed toward more robust speech representations, a key challenge emerged: developing DL models that can identify stuttering related characteristics despite variations across speakers, languages and acoustic conditions. To address this challenge, recent approaches have incorporated speaker metadata and multi-task learning, including adversarial learning techniques that suppress speaker specific information to promote representations that generalize across different stuttering types~\citep{sheikh2022robust,liu2025addressing}. Building on these advances in representation learning, transformer based architectures such as wav2vec~2.0~\citep{baevski2020wav2vec}, which rely on self attention mechanisms to model long range interactions~\citep{vaswani2017attention}, have further improved the ability to capture contextual information in speech and have achieved state of the art performance in automated stuttering detection~\citep{sheikh2022mlstuttering, sheikh2022endoend, schuller2022acm}. At the same time, benchmark initiatives such as the \emph{Stuttering Sub-Challenge} of the \emph{Computational Paralinguistics Challenge} (ComParE) have provided standardized datasets and evaluation protocols, enabling systematic comparison of competing approaches and accelerating methodological progress~\citep{schuller2022acm,bayerl2023classification}. Together, these developments have transformed automated stuttering detection from a feature engineering problem into an advanced computational speech analysis task and have established it as a potentially useful source of information for clinical decision support. Nevertheless, accurate automatic stuttering detection alone does not address the subsequent and clinically more consequential question of how such information should inform intervention planning. This distinction motivates the transition from automated stuttering speech analysis towards systems that can assist with the broader therapeutic workflow, while retaining appropriate clinician oversight.

\section{Agentic AI: A Clinician-in-the-Loop Approach}
The growing capability of automated stuttering speech analysis models therefore provides an important technological foundation, but it represents only one stage of the therapeutic process. Once information about a person's speech has been extracted, this information must still be interpreted in the context of the individual's communication goals, clinical presentation, and broader psychosocial circumstances before it can meaningfully inform intervention. Existing digital speech therapy tools have largely focused on structured practice, monitoring, or telepractice support, rather than dynamically integrating speech derived information into an adaptive therapeutic planning process. Consequently, the challenge is no longer only how to detect stuttering and its types, but also how systems might use such information to support the therapeutic planning that follow assessment.

This broader challenge has contributed to growing interest in \emph{agentic AI}, in which systems are designed not merely to produce isolated predictions or responses, but to perceive relevant information, reason over that information, and iteratively generate or execute actions toward a defined objective~\citep{karunanayake2025next,gridach2026agentic}. In a therapeutic context, such capabilities could allow an AI system to combine speech derived information $\mathcal{S}$ with contextual and clinical inputs when proposing candidate intervention strategies. However, the ability to generate a therapeutic recommendation does not imply that the system should independently determine or deliver clinical care. Instead, a more appropriate framework is a \textit{clinician-in-the-loop} (CITL) paradigm, in which AI-generated therapy plans or recommendations remain subject to human review, interpretation, and decision making~\citep{tang2020clinician,zou2025rise}. Such a division of labor allows computational systems to assist with information processing and candidate therapy plan generation while preserving the SLP's role in clinical reasoning, contextual interpretation, and final decision making.

This distinction is particularly important in speech therapy because successful intervention depends not only on observable speech characteristics but also on individual goals, communicative demands, emotional experiences, motivation, and the therapeutic relationship~\citep{guitar2018stuttering}. An agentic system could therefore support aspects of therapy planning by organizing speech derived information $\mathcal{S}$, proposing candidate exercises, and adapting suggestions to available contextual information, while the SLP evaluates whether those suggestions are appropriate for the individual or requires any further modifications~\citep{kim2025clinical}. The same principle could potentially support longitudinal care; speech samples collected across sessions might provide additional information that helps a clinician monitor changes and identify issues requiring attention between appointments. However, the current VST system analyzes the speech sample submitted with each request, whereas the incorporation of longitudinal speech information, which would require fine-tuning foundation models, has not been evaluated and therefore remains outside the scope of the current work.

Although agentic AI is increasingly being investigated across medical domains~\citep{karunanayake2025next, powell2025agentic, dietrich2025agentic,abdollahi2026next}, its application to speech and language therapy remains limited. Recent studies have explored LLMs for intervention planning~\citep{kim2025clinical, birol2025there} and for answering questions related to stuttering~\citep{saeedi2025assessing}, while other work has investigated LLMs for stuttering detection or assistive applications~\citep{wagner2024large, faggiani2025easetalk, aljameel2025stuttering, jokar2026advancing}. Related developments in other speech disorders such as aphasia have also begun to explore therapist supervised LLM agent workflows~\citep{galindo2026rehabilitia}. These studies collectively demonstrate growing interest in applying AI to different components of speech and language care, but they also highlight a separation between automated speech analysis, LLM based reasoning, and clinician supervised therapeutic planning. \textcolor{black}{To our knowledge, no clinician supervised multi-agent LLM system} specifically designed for stuttering therapy planning has yet been reported, representing a gap that motivates the present work~\citep{sheikh2025overview}. This gap is particularly relevant because stuttering is heterogeneous in both its observable speech characteristics and its individual and psychosocial impact, making clinical interpretation and individualized decision making essential~\citep{jokar2026addressing}. A CITL agentic framework can therefore be viewed not as an attempt to replace the SLP, but as a means of connecting automated speech analysis including stuttering detection $f_{SD}$ with AI-assisted therapeutic planning while keeping clinical judgment at the center of the process. In brief, we summarize our contributions as follows:
\begin{itemize}
\item \textbf{Virtual Speech Therapist:} We present VST, a proof-of-concept framework that integrates self-supervised stuttering detection $f_{SD}$, agentic therapy planning, and a clinician supervised loop for reviewing, modifying, and approving candidate therapy plans.

\item \textbf{Multi-agentic therapy planning:} We develop an LLM based multi-agentic system in which a \textit{TherapyAgent} $\pi_{ta}$ generates candidate therapy plans $\mathcal{P}$ using speech derived information $\mathcal{S}$ and evidence based protocols, while a \textit{CriticAgent} $\pi_{ca}$ provides structured feedback $\mathcal{R}$ and acts as \textit{proxy} SLP to support subsequent plan refinement.

\item \textbf{CITL decision support:} We position VST as a decision support system in which the clinician retains responsibility for interpreting the generated information, refining therapeutic goals, and approving the final plan $\mathcal{P}$, rather than delegating clinical decision making to the AI system.
\end{itemize}
}

\begin{figure*}[!t]
  \centering
    \includegraphics[scale=0.86]{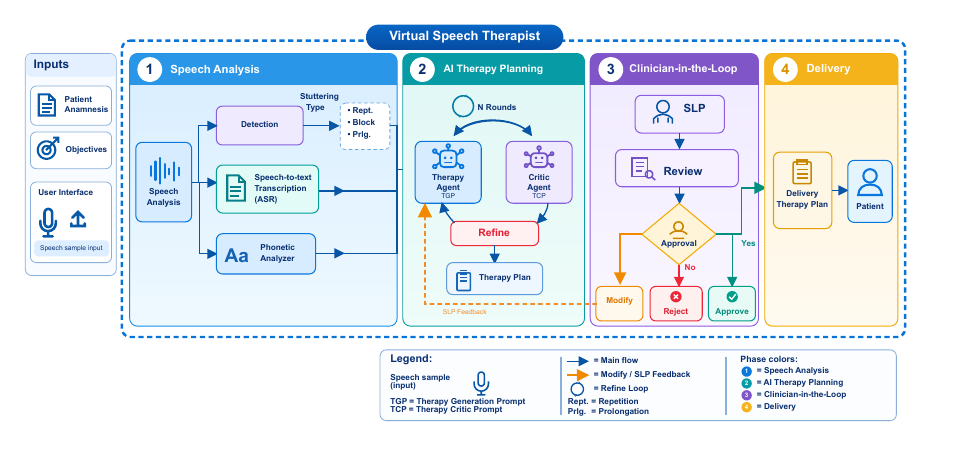}
  \caption{End-to-end pipeline of the Virtual Speech Therapist (VST). A stuttered speech sample $\textbf{x}$ is analyzed in parallel by the stuttering detection module $f_{SD}$, the automatic speech recognition module $f_{ASR}$, and the phonemization module $f_{ph}$. The $f_{SD}$ module identifies stuttering types $\mathbf{y}_{st}$, including \textit{repetitions, blocks, and prolongations}, while the $f_{ASR}$ and $f_{ph}$ modules generate the textual transcription $\mathbf{T}$ and phonetic representation $\mathbf{\Phi}$, respectively. These outputs are then used by the VST to generate a proposed therapy plan $\mathcal{P}$, which is submitted to an SLP for clinical review (\Cref{eq:mod}). Based on this review, the SLP provides feedback $\mathcal{R}$ and may request revisions to the proposed plan. The SLP retains final decision making authority (\Cref{eq:dec,eq:mod}) to approve or reject the plan or request further modifications, thereby ensuring that the therapy plan $\mathcal{P}$ is not finalized without explicit clinician oversight.}
  \label{fig:pipeline}
\end{figure*}

\section{Proposed System: Virtual Speech Therapist}
\label{sec:methods}
{
\color{black}
To investigate how automated speech information $\mathcal{S}$ can be incorporated into clinician supervised therapy planning, we developed a VST prototype (\Cref{fig:pipeline}) that links stuttering detection $f_{SD}$ with an LLM based, multi-agentic therapy planning workflow. Rather than treating the output of automated detection as a clinical decision, the VST uses it as one source of structured information $(\mathcal{S})$ from which candidate therapy plans $(\mathcal{P})$ are generated and subsequently reviewed by an SLP (\Cref{eq:dec,eq:mod}). The system therefore follows a CITL design framework in which automated components support information extraction, plan generation, and iterative refinement~\citep{madaan2023self}, while responsibility for \textit{clinical interpretation and intervention} remains with the clinician. In accordance with this design, a generated plan $\mathcal{P}$ is not intended for delivery to a client until it has been reviewed and approved by an SLP, and implementation of the approved plan remains under clinical supervision. Unsupervised use by clients is outside the scope of the present prototype. 

The resulting VST system is organized into following sequential phases that reflect the progression from speech observation to clinician supervised intervention: (1) speech processing and stuttering detection; (2) candidate therapy planning and structured critique; encompassing prompt designs, the VST agentic workflow with iterative refinement, and CITL validation.

\subsection{Speech Pre-processing and Stuttering Detection}
\label{sec:sub:sp_sd}
The first phase converts the given stuttered speech sample $\mathbf{x}$ into structured information $\mathcal{S}$ that can be used by the VST therapy planning module. This phase is designed to preserve the link between the original speech sample and the derived information $\mathcal{S}$ subsequently provided to the LLM agents, rather than relying solely on manually entered clinical descriptions. It consisted of two sequential steps: speech sample preparation and automated stuttering detection.
 \begin{itemize}
\item {\textit{Preprocessing}.}
The VST first receives a speech sample $\mathbf{x}$ through its user interface\footnote{\href{https://www.vocametrix.com/ai/stuttering-therapy-planning-agent}{https://www.vocametrix.com/ai/stuttering-therapy-planning-agent}}, which supports either direct audio recording or upload of a previously recorded speech file. Once submitted, the speech sample $\mathbf{x}$ is preprocessed and segmented into overlapping speech windows using configurable window and overlap durations (\Cref{tab:segmentation_params}). This segmentation converts the continuous speech into temporal units that can be analyzed consistently by the downstream stuttering detection model $f_{SD}$ while preserving local acoustic context. The resulting speech segments are then passed to the automated speech analysis stage.

\item {\textit{Detection}.}
Each speech segment is subsequently passed to the trained stuttering detection model $f_{SD}$(Section~\ref{sec:sd}) (\Cref{eq:sd}), which identifies the given speech segment according to the detected stuttering type $\mathbf{y_{st}}$ as a \emph{block}, \emph{prolongation}, or \emph{repetition}, etc. In parallel, an automatic speech recognition $f_{ASR}$ module (\Cref{eq:asr}) transcribes each stuttered speech segment to capture its linguistic content $\mathbf{T}$, while a phonetic analysis module $f_{ph}$ (\Cref{eq:ph}) derives its corresponding phonetic representation $\mathbf{\Phi}$. The detected stuttering information and ASR transcription are subsequently combined, thereby linking the observed stuttering characteristics with the content of the corresponding utterance. This integrated representation provides the speech derived information $\mathcal{S}$ (\Cref{eq:str_info}) that is passed to the VST therapy planning module as the basis for generating candidate intervention plans.
\begin{tcolorbox}[
    colback=cyan!5!white,
    colframe=blue!65!black,
    coltext=blue!50!black,
    boxrule=1pt,
    arc=3mm,
    left=3mm,
    right=3mm,
    top=2mm,
    bottom=2mm,
    breakable
]
Mathematically, we can formulate this as: 
\begin{align}
    \mathbf{y}_{\text{st}} &= f_{\text{SD}}(x),
    \label{eq:sd}\\
    \mathbf{T} &= f_{\text{ASR}}(x),
    \label{eq:asr}\\
    \mathbf{\Phi} &= f_{\text{ph}}(x),
    \label{eq:ph}\\
    \mathcal{S} &= \left[ \mathbf{y}_{\text{st}}, \mathbf{T}, \mathbf{\Phi} \right],
    \label{eq:str_info}
\end{align}
where $\textbf{x}$ is the input stuttered speech sample and $\mathcal{S}$ represents the complete structured stuttering profile,
combining the predicted stuttering type ($\mathbf{y}_{\text{st}}$), textual transcription ($\mathbf{T}$),
and phonetic representation ($\mathbf{\Phi}$). This profile is subsequently provided
as input to the therapy planning module.

\end{tcolorbox}
\end{itemize}
Thus, the first phase establishes a continuous pathway from the given stuttered speech sample $\mathbf{x}$ to a structured representation $\mathcal{S}$ that captures both \emph{how} the person speaks, through detected stuttering characteristics $\mathbf{y}_{st}$, and \emph{what} was spoken, through the corresponding transcription ($\mathbf{T}$, $\mathbf{\Phi}$). This representation serves as the input to the next phase, where the VST uses this speech derived information $\mathcal{S}$ together with relevant therapeutic context and evidence based protocols (embedded in prompts, cf. \Cref{sec:prompts}) to generate and iteratively refine candidate therapy plans $\mathcal{P}$.}

{
\color{black}
\subsection{Prompts and Therapy Planning via Iterative Refinement}
\label{sec:agents}

Following the detection phase described in \Cref{sec:sub:sp_sd}, the VST system (\Cref{fig:pipeline}) initiates an agentic orchestration workflow in which LLM based agents collaborate to generate, evaluate, and iteratively refine therapy plans $\mathcal{P}$. Inspired by the self-refinement mechanism proposed by \citet{madaan2023self}, this workflow incorporates an iterative critique process intended to act as a \textit{proxy} for SLP review, with the aim of improving the consistency and clinical soundness of the proposed therapy plans. 
}
\subsubsection{Prompt Designs}
\label{sec:prompts}
In this section, we briefly describe the prompt design choices for the LLM agents, which were developed through consultation with an SLP (G.L) and a phonetician (F.H).

{
\color{black}
\textbf{Therapy Generation Prompt}: Because the behavior of an LLM based VST agent depends substantially on how the clinical task and its constraints are specified, prompt design was treated as part of the methodological formulation rather than as an incidental implementation detail. The \textit{Therapy Generation Prompt} (TGP) prompt (denoted as \(\mathcal{I}_{\text{TGP}}\)) was therefore developed iteratively using input from SLP and a phonetician together with qualitative inspection of outputs from representative speech samples and stuttering profiles. Its design incorporated four principal constraints: (i) adherence to evidence based speech therapy practices, (ii) use of supportive and non-judgmental language, (iii) adaptation of candidate strategies to the detected stuttering characteristics and linguistic context, and (iv) avoidance of medical diagnosis or clinical claims beyond the intended scope of the system. These constraints were introduced to reduce the likelihood of inappropriate or unsupported recommendations, although they cannot by themselves guarantee the clinical appropriateness of an output.
\par 
To determine whether the resulting \(\mathcal{I}_{\text{TGP}}\) prompt provided sufficiently relevant and consistent responses for the intended workflow, multiple candidate \(\mathcal{I}_{\text{TGP}}\) prompt formulations were examined using representative speech samples and stuttering profiles. The comparisons considered response relevance, therapeutic clarity, linguistic accuracy, and consistency across outputs. The final \(\mathcal{I}_{\text{TGP}}\) prompt was selected based on informal qualitative comparisons by the authors, in which it produced the most relevant and consistent responses. However, no formal prompt ablation or quantitative evaluation was conducted. The complete \(\mathcal{I}_{\text{TGP}}\) prompt is provided in \textit{Appendix}~\ref{lab:appendix} to make the assumptions and constraints governing the $\pi_{ta}$ agent explicit and reproducible.
}

{
\color{black}
\textbf{Therapeutic Knowledge Encoding}: The  \(\mathcal{I}_{\text{TGP}}\) prompt further incorporates the established therapeutic principles by instructing the $\pi_{ta}$ agent to consider four broad families of approaches: {\textit{stuttering modification}, including voluntary stuttering, pull-outs, and cancellations~\citep{vanriper1973treatment}; \textit{fluency shaping}, including easy onsets, light articulatory contacts, and continuous phonation; \textit{avoidance reduction therapy}, including desensitization and fear reduction strategies~\citep{sisskin2022avoidance}; and \textit{cognitive behavioral approaches}, including cognitive behavioural therapy (CBT), acceptance and commitment therapy (ACT), and rational emotive behaviour therapy (REBT)}. These approaches were incorporated because they address different dimensions of the speaking experience. \textit{Fluency shaping} emphasizes increased fluency, whereas \textit{stuttering modification} emphasizes reducing struggle and increasing ease during moments of stuttering, with integrated approaches combining elements of both~\citep{guitar2018stuttering}. These principles guide candidate plan $\mathcal{P}$ generation by linking the output of the trained stuttering detection model $f_{SD}$ to the corresponding therapeutic strategies. For example, detected repetitions may guide the generation of fluency shaping tasks, while detected blocks may inform stuttering modification tasks; when multiple stuttering patterns are identified, the generated plan $\mathcal{P}$ may incorporate a combination of strategies. Avoidance reduction and cognitive strategies are incorporated alongside these speech focused techniques, and the generated plan is structured to progress from awareness and isolated practice toward words and phrases, sentences, and increasingly realistic speaking contexts, with measurable goals and self-monitoring scales. Importantly, these mappings represent instructions provided to the VST system rather than findings established by the present study; the appropriateness of a particular therapeutic approach for an individual remains a matter of clinical judgment. The corresponding instructions are therefore documented in Appendix~\ref{lab:appendix} rather than treated as an autonomous clinical decision.
}

{
\color{black}
\textbf{Therapy Critic Prompt:}~To provide a structured basis for reviewing the generated candidate therapy plan \(\mathcal{P}\), we formulated a dedicated \textit{Therapy Critic Prompt} (TCP) (denoted as \(\mathcal{I}_{\text{TCP}}\)), analogous to the \(\mathcal{I}_{\text{TGP}}\) prompt, which specifies the criteria against which the generated plan \(\mathcal{P}\) should be examined by the $\pi_{ca}$ agent. The \(\mathcal{I}_{\text{TCP}}\) prompt defines six complementary review domains. First, \emph{clinical soundness} considers whether the proposed interventions are aligned with the detected stuttering type and severity. Second, \emph{safety and psychological impact} considers potential risks associated with physical strain, vocal misuse, or adverse psychological effects. Third, \emph{evidence alignment} examines whether the proposed strategies are consistent with peer-reviewed evidence and established clinical guidance, including resources such as American speech language hearing association (ASHA) and international fluency association (IFA). Fourth, \emph{structure, clarity, and practicality} addresses the organization, readability, feasibility, and potential usability of the plan by an SLP. Fifth, \emph{improvements and missing elements} directs the review toward identifying gaps or potential modifications in intervention sequencing, intensity, or therapeutic coverage. Finally, \emph{explainability and reasoning transparency} requires each proposed strategy to be supported by a structured \emph{clinical reasoning} chain specifying the underlying observation, rationale, expected outcome, and evidence base. The complete \(\mathcal{I}_{\text{TCP}}\) is provided in \textit{Appendix}~\ref{lab:appendix}.
}


{
\color{black}
\subsubsection{VST Architecture}
\label{sec:vst_agent}
This section describes the components of the VST system. We first introduce the primary generative and evaluative modules, the \textit{TherapyAgent} $\pi_{ta}$ and \textit{CriticAgent} $\pi_{ca}$, followed by their iterative self-refinement step and the mechanisms enabling real time CITL feedback.

\textbf{TherapyAgent~($\pi_{ta}$)}: The speech derived information $\mathcal{S}$ established in \Cref{sec:sub:sp_sd}, including the detected stuttering characteristics $\mathbf{y}_{st}$, is provided to the $\pi_{ta}$ agent together with the $\mathcal{I}_{\text{TGP}}$ prompt, which specifies the clinical constraints and therapeutic principles described above. The $\pi_{ta}$ agent uses this information to generate (\Cref{eq:initial_therapy_plan}) a candidate therapy plan $\mathcal{P}$ that translates the observed speech characteristics into a structured set of potential interventions, including fluency shaping exercises, cognitive behavioral tasks, and a progression schedule. The $\pi_{ta}$ agent therefore serves as the primary plan generation component of the VST workflow, transforming speech derived information $\mathcal{S}$ and predefined therapeutic guidance into a candidate intervention plan $\mathcal{P}$ for subsequent computational critique and clinician review. The $\pi_{ta}$ agent is based on LLM from the \textit{Gemini} model family (cf. \Cref{sec:llms}). The underlying LLM provides the language based reasoning and generation capabilities required to interpret the structured input from Phase~1, incorporate the constraints encoded in the $\mathcal{I}_{\text{TGP}}$, and formulate the candidate therapy plan $\mathcal{P}$. The LLM itself is not treated as a clinical decision maker, rather, it functions as the generative reasoning engine within the broader CITL framework. The resulting plan $\mathcal{P}$ is consequently treated as a candidate output that requires subsequent finetuning and SLP review before it can be considered for clinical use. 
\begin{tcolorbox}[
    colback=cyan!5!white,
    colframe=blue!65!black,
    coltext=blue!50!black,
    boxrule=1pt,
    arc=3mm,
    left=3mm,
    right=3mm,
    top=2mm,
    bottom=2mm,
    breakable
]
\textit{Mathematically,}:
\begin{equation}
    \mathcal{P}_0 \sim
    \pi_{ta}\left(
        \mathcal{S}, \mathcal{I}_{\text{TGP} } 
    \right) 
    \label{eq:initial_therapy_plan}
\end{equation}
\textit{where} $\mathcal{S}$ is stuttering profile, $\mathcal{I}_\text{TGP}$ is therapy generation prompt, and $\mathcal{P}_0$ \textit{denotes the initial candidate therapy plan}.
\end{tcolorbox}

\textbf{CriticAgent~($\pi_{ca}$)}: The $\pi_{ca}$ agent is implemented using the same family of LLMs as the $\pi_{ta}$ agent, but with a different hyperparameter settings to support its distinct critique oriented role. Rather than generating a therapy plan $\mathcal{P}$ independently, the $\pi_{ca}$ agent serves as an SLP informed proxy reviewer that examines the candidate plan $\mathcal{P}$ produced by the $\pi_{ta}$ agent (\Cref{eq:critic}). Guided by the six review domains specified in the $\mathcal{I}_{\text{TCP}}$ prompt, it uses its LLM based architecture (\Cref{sec:llms}) to assess the candidate plan $\mathcal{P}$ and identify potential inconsistencies, omissions, or areas requiring further refinement. 
\begin{tcolorbox}[
    colback=cyan!5!white,
    colframe=blue!65!black,
    coltext=blue!50!black,
    boxrule=1pt,
    arc=3mm,
    left=3mm,
    right=3mm,
    top=2mm,
    bottom=2mm,
    breakable
]
Mathematically, we can formulate the critique generation as: 
\begin{align}
    C_\mathcal{N} &\sim \pi_{ca}\left(
        \mathcal{S}, \mathcal{P}_{\mathcal{N}-1},
        \mathcal{I}_{\text{TCP}}
    \right)
    && \text{(Critic Evaluation)}
    \label{eq:critic}
\end{align}
\textit{Here},  $\mathcal{S}$ is stuttering profile, $\mathcal{I}_\text{TCP}$ is therapy critic prompt, $\mathcal{P}_{\mathcal{N}-1}$ is the therapy plan generated at previous round, and $C_\mathcal{N}$ \textit{denotes the critic feedback at round} $\mathcal{N}$.
\end{tcolorbox}

\textbf{Iterative Refinement:} The resulting structured feedback is then provided back to the $\pi_{ta}$ agent to guide subsequent revision (\Cref{eq:refine}), thereby establishing a generation-critique-refinement~\citep{madaan2023self} cycle while preserving the SLP as the final clinical decision maker. For each domain, the $\pi_{ca}$ agent produces structured observations, identifies strengths and potential concerns, and provides actionable recommendations for revision. When potential discrepancies, inconsistencies, or suboptimal therapeutic choices are identified, the $\pi_{ca}$ agent explicitly flags the affected components and provides feedback that can be passed back to the $\pi_{ta}$ agent. Depending on the identified issue, this feedback may suggest alternative therapeutic strategies, modifications to intervention sequencing or intensity, or consideration of relevant clinical protocols and empirical evidence. In this workflow, the $\pi_{ca}$ agent therefore functions as a structured computational \textit{proxy} SLP that facilitates iterative plan refinement. Importantly, its output represents LLM based self-review rather than independent clinical validation; the resulting critique is intended to identify potential issues for subsequent refinement and does not replace the SLP's clinical assessment. This iterative design was motivated by the observation that a single generation step provides no explicit mechanism for responding to identified weaknesses; the refine loop instead allows the system to incorporate its own structured feedback before the output reaches the clinician~\citep{madaan2023self}. The loop is executed for a predefined number of rounds, $\mathcal{N}$, with a default of $\mathcal{N}=2$ and a configurable maximum of five rounds in the user interface. After the prescribed rounds, the resulting plan $\mathcal{P}$ is treated as a refined candidate rather than as a clinically validated intervention and is forwarded to the SLP for review. 
\begin{tcolorbox}[
    colback=cyan!5!white,
    colframe=blue!65!black,
    coltext=blue!50!black,
    boxrule=1pt,
    arc=3mm,
    left=3mm,
    right=3mm,
    top=2mm,
    bottom=2mm,
    breakable
]
Mathematically, we formulize the iterative refinement as: 
\begin{align}
    \mathcal{P}_\mathcal{N} &\sim \pi_{ta}\left(
        \mathcal{S}, \mathcal{P}_{\mathcal{N}-1},
        C_\mathcal{N}, \mathcal{I}_{\text{TGP}}
    \right)
    && \text{(Iterative Refinement)}.
    \label{eq:refine}
\end{align}
\textit{The revised plan $\mathcal{P}_\mathcal{N}$ is generated conditionally on the
original stuttering profile ($\mathcal{S}$), the previous therapy plan generated ($\mathcal{P}_{\mathcal{N}-1}$), and the ($\pi_{ca}$) agent's
feedback ($\mathcal{C}_{\mathcal{N}}$). After $\mathcal{N}$ refinement rounds, $\mathcal{P}_\mathcal{N}$ is obtained
as the final candidate plan for subsequent review by an SLP}.
\end{tcolorbox}

\textbf{CITL Validation:}~The transition from computational refinement to the SLP review constitutes the central boundary of the VST framework. Although the $\pi_{ta}$ agent and $\pi_{ca}$ agent can generate and critique candidate plans $\mathcal{P}$, neither component is assumed to possess sufficient clinical context or authority to determine whether a plan $\mathcal{P}$ should be delivered to a patient. The refined plan $\mathcal{P}$ is therefore presented to the SLP, who evaluates its clinical appropriateness, patient safety, consistency with evidence based practice, clarity, progression, feasibility, and fit with the individual's fluency profile. In this way, the clinician does not merely perform a final technical check, rather, the SLP provides the clinical interpretation that the VST workflow is explicitly designed not to replace. The clinician can consequently make a decision $\mathcal{D_{SLP}}$ (\Cref{eq:dec}) either to approve the plan $\mathcal{P}$, reject it and terminate the VST workflow, or request specific modifications, for example to the intervention type, intensity, or sequencing. When modifications are requested, the clinician's feedback $\mathcal{R}$ is returned to the $\pi_{ta}$ agent for one additional refinement round (\Cref{eq:mod}), after which the revised final therapy plan $\mathcal{P}_{\mathcal{N}+1}$ is returned back to the SLP for a further decision. This final human guided loop establishes the CITL paradigm as the organizing principle of the VST workflow; automated speech analysis provides the speech derived evidence, the prompts translate this information into clinically constrained inputs for the agents, the agents generate and critique a candidate therapy plan $\mathcal{P}$, and the clinician ultimately determines whether and how the plan should proceed. Thus, the role of agentic AI in VST is deliberately bounded by the clinical decision making process. The generated therapy plan $\mathcal{P}$ is a LLM driven produced draft whose purpose is to support, rather than replace, clinician reasoning. No plan is considered ready for delivery without an explicit clinician decision ($\mathcal{D}_{SLP}$), and any subsequent implementation remains under clinical supervision. Accordingly, the present study focuses on the feasibility of the proposed VST based therapy planning workflow rather than its clinical deployment. To illustrate the outputs produced by this workflow, representative therapy plans $\mathcal{P}$ generated by the VST system are provided in \textit{Appendix}~\ref{lab:appendixb}.

}
\begin{tcolorbox}[
    colback=cyan!5!white,
    colframe=blue!65!black,
    coltext=blue!50!black,
    boxrule=1pt,
    arc=3mm,
    left=3mm,
    right=3mm,
    top=2mm,
    bottom=2mm,
    breakable
]
\textit{Mathematically, the CITL validation can be formularized as:
\begin{align}
&\mathcal{D}_{\mathrm{SLP}}\left(
    \mathcal{P}_{\mathcal{N}}, \mathcal{R}_{\mathcal{N}}
\right)
\in \{\text{approve},\text{reject},\text{revise}\},
\label{eq:dec}\\
&\mathcal{P}_{\mathcal{N}+1}
\sim \pi_{ta}\left(
    \mathcal{S}, \mathcal{P}_{\mathcal{N}},
    \mathcal{R}_{\mathcal{N}}, \mathcal{I}_{\mathrm{TGP}}
\right),
\quad \text{if } \mathcal{D}_{\mathrm{SLP}}=\text{revise},
\label{eq:mod}
\end{align}
$\mathcal{D}_{SLP}$ denotes the
SLP's decision at review round $\mathcal{N}$, and $\mathcal{R}_\mathcal{N}$ represents the
requested revisions. The revised plan is resubmitted to the SLP,
and the loop continues until the plan is approved or rejected.}
\end{tcolorbox}


\section{Experimental Setup}
\label{Section:ExperimentalSetup}
\textbf{Dataset:}~We utilize the SEP-28k-Extended (SEP-28k-E) dataset~\citep{bayerl_sep28k_E_2022}, an extension of the SEP-28k stuttering dataset\footnote{The largest publicly available dataset for stuttering detection.}~\citep{lea:2021}. The dataset comprises approximately 28k three second audio clips sourced from podcasts, annotated by three trained non-clinicians. The annotations use a multi-label scheme comprising five stuttering categories: \textit{prolongation, block, sound repetition, word repetition, and interjection}. We used this dataset to train the stuttering detection model integrated into the VST. To assess the subsequent VST therapy planning workflow, we evaluated VST using 16 speech recordings (prolongations, repetitions (sound and word), blocks and interjection) from adults that had been collected previously during routine clinical practice at the author's (G.L) clinic and were anonymized before use.  

\label{sec:embeds}
\par

{
\color{black}
\subsection{Stuttering Detection}
\label{sec:sd}

Within VST system, stuttering detection $f_{SD}$ provides the first computational module of the given stuttered speech sample $\textbf{x}$. The objective at this stage is not to make a clinical diagnosis, but to identify the observable stuttering events $\mathbf{y}_{st}$ in the given audio recording that can subsequently be used as one source of information for VST therapy planning. To obtain these speech derived labels, we investigate two complementary methods based on the pretrained wav2vec~2.0 (XLSR-53) model~\citep{conneau2020xlsr53}. The first treats XLSR-53 as a fixed representation learner and uses its contextual embeddings as input to a lightweight downstream classifier, whereas the second adapts the complete pretrained model to the stuttering detection task through end-to-end fine-tuning. This comparison allowed us to examine whether task specific adaptation provides an advantage over using the pretrained representations directly, while keeping the role of detection clearly separated from the primary contribution of this study, namely the VST\footnote{\textit{It is important to emphasize that enhancing stuttering detection accuracy is not the primary focus of this study. We assume the availability of state-of-the-art detection models that already achieve competitive performance. Rather, the principal contribution of this work is the design of an agentic framework for automated and adaptive therapy planning within the \textit{Virtual Speech Therapist} system.}}. The stuttering detection module operates exclusively on the acoustic signal and receives 3--5~s speech windows represented using XLSR-53 embeddings. Neither the ASR transcription $\mathbf{T}$ nor the phonemic description $\mathbf{\Phi}$ are used by the classifier $f_{SD}$. Instead, these linguistic representations are generated separately and introduced at the therapy planning stage, where they provide additional context for the $\pi_{ta}$ agent.

The stuttering detection model $f_{SD}$ is trained to identify the five audio based event types annotated in SEP-28k-E: \textit{prolongation, block, sound repetition, word repetition, and interjection}. Because these labels are derived from audible speech events, the current detection model does not capture silent blocks with no acoustic correlate or secondary behaviors that are visible but not represented in the audio, such as head movements, eye blinking, or facial tension. Blocks also remain the most challenging category for the current model, with an F1 score of 32.0\% (\Cref{tab:f1_comparison}). These limitations are relevant when interpreting the information subsequently supplied to the therapy planning agents and reinforce that the detector provides partial speech derived evidence rather than a complete clinical assessment. To make this uncertainty visible to the clinician, the classifier confidence is additionally summarized as {high}, {medium}, or {low} using thresholds of 0.8 and 0.6, respectively. Importantly, this confidence label does not prevent a recording from proceeding to therapy plan generation.

\textbf{Shallow Neural Network:}~ Our first approach uses the pretrained multilingual wav2vec~2.0 XLSR-53 model as a feature extractor. Developed by Meta and pretrained on approximately 50,000 hours of speech spanning 53 languages, XLSR-53 produces contextual representations that capture a range of speech related and paralinguistic features~\citep{sheikh2022endoend, bayerl2023classification}. For each speech sample $\mathbf{x}$, we extract these representations and apply statistical pooling to convert the variable-length embedding sequence into a fixed dimensional 1024-dimensional feature vector. This vector is then passed to a lightweight three layer neural network classification head to identify stuttering related speech patterns~\footnote{\url{https://github.com/shakeel608/stutternet/tree/main}}. The classifier contains two hidden layers of 64 units with exponential linear unit (ELU) activations, followed by a fully connected output layer that maps the learned representation to logits for the target stuttering event classes and their corresponding probabilities. This design builds on established approaches that use pretrained contextual speech representations for paralinguistic and stuttering analysis~\citep{bayerl2023classification,bayerl22b_interspeech,sheikh2022introducing,sheikh2023stuttering,sheikh2022endoend,sheikh2023deep}, while keeping the downstream classifier sufficiently lightweight for local inference.

\textbf{End-to-end wav2vec~2.0 XLSR-53 fine-tuning:} ~As a second strategy, we investigated whether adapting the pretrained acoustic representation directly to the stuttering detection task could provide additional discriminative information. We replaced the original head of the pretrained wav2vec~2.0 XLSR-53 checkpoint with a stuttering task specific classification head and fine-tuned the entire model on the SEP-28k-E dataset using the validated speaker disjoint data split~\citep{bayerl_sep28k_E_2022}. During fine-tuning, the model parameters were optimized using the cross entropy loss defined in Equation~\eqref{eq:cross_entropy}. Compared with the frozen feature configuration, fine-tuning produced improved performance across most stuttering categories, with particularly notable gains for sound repetition, word repetition, and interjection. A modest improvement was also observed for the challenging block category, while performance on prolongation decreased slightly, as shown in \Cref{tab:f1_comparison}. 

\begin{tcolorbox}[
    colback=cyan!5!white,
    colframe=blue!65!black,
    coltext=blue!50!black,
    boxrule=1pt,
    arc=3mm,
    left=3mm,
    right=3mm,
    top=2mm,
    bottom=2mm,
    breakable
]
\begin{equation}
\mathcal{L}_{\mathrm{CE}}
=
-\frac{1}{N}
\sum_{i=1}^{N}
\sum_{c=1}^{C}
y_{i,c}\log\left(\hat{p}_{i,c}\right),
\label{eq:cross_entropy}
\end{equation}

where \(N\) is the number of training samples, \(C\) is the number of stuttering classes, \(y_{i,c}\) is the ground truth annotation, and \(\hat{p}_{i,c}\) is the predicted probability for class \(c\), obtained using the softmax function.
\end{tcolorbox}

\textbf{Role of detection within the VST pipeline:}~The two detection $f_{SD}$ strategies therefore represent alternative implementations of the same upstream function within VST framework; converting the acoustic speech signal into structured stuttering information $\mathcal{S}$. This information $\mathcal{S}$ is subsequently combined with the independently generated $\mathbf{T}$ and $\mathbf{\Phi}$ transcriptions and passed to the $\pi_{ta}$ agent (cf. \Cref{sec:vst_agent}). In this way, the detection component $f_{SD}$ does not constitute the endpoint of the proposed system; rather, it establishes the stuttering evidence $\mathbf{y}_{st}$ on which the subsequent agentic therapy planning process operates. The distinction is central to the scope of the present study; although we evaluate two approaches to speech level stuttering classification, the principal research focus is how this information can be incorporated into a CITL workflow for generating, critiquing, and refining candidate therapy plans $\mathcal{P}$.
}

\begin{table}[!ht]
  \centering
  \caption{Comparative F1 Scores with fine tuning (FT) XLRS53 and without finetuning (\%)}
  \label{tab:f1_comparison}

  \begin{tabular}{lcc}
    \toprule
    \textbf{Class} & {$\text{\textbf{Fine tuning}}$} & {$\text{\textbf{SOTA (No FT)}}$} \\
    \midrule
    Soundrepetition & {\bfseries {43.00}} & 32.07 \\
    Wordrepetition  & {\bfseries {56.00}} & 41.23 \\
    Block           & {\bfseries {32.00}} & 31.02 \\
    Fluent          & {\bfseries {82.00}} & 66.92 \\ 
    Interjection    & {\bfseries {77.00}} & 51.63 \\
    Prolongation    & 44.00 & {\bfseries {46.23}} \\
    \midrule
    \textbf{Weighted Average F1} & {\bfseries {67.00}} & 44.85 \\
    \bottomrule
  \end{tabular}
\end{table}

{
\color{black}
\subsection{LLM Backbone of the VST Agents}
\label{sec:llms}

The agentic therapy planning workflow described in Section~\ref{sec:vst_agent} requires an LLM capable of interpreting the structured speech derived information $\mathcal{S}$ from \Cref{sec:sub:sp_sd}, incorporating the therapeutic constraints encoded in the $\mathcal{I}_{TGP}$ prompt, generating a coherent candidate therapy plan $\mathcal{P}$ via $\pi_{ta}$ agent, and subsequently examining that plan through structured critique via $\pi_{ca}$ agent. The choice of the underlying LLM is therefore an important component of the VST system, as the same language generation capability supports both plan generation and computational review. Although several foundational LLMs have demonstrated strong performance across natural language processing and speech related tasks~\citep{9893562}, including OpenAI's GPT, DeepSeek, Mistral, Anthropic's Claude, and Google's Gemini, evaluating multiple model families was beyond the scope of this study. This decision was primarily driven by limited funding, as many state of the art frontier models require paid inference services with varying costs. We therefore selected a single model family for both agents in VST system to maintain methodological consistency across the therapy generation and critique stages while keeping computational costs manageable.

Based on preliminary comparisons of therapeutic content generation, we selected \textit{Gemini 3 Pro (Preview)}~\citep{comanici2025gemini} as the underlying LLM for both agents. The selection was guided by the model's ability to handle long contextual inputs, its reasoning capabilities, and its inference efficiency, which together provided a suitable balance for the requirements of the VST workflow. Using the same model family for both agents also allowed us to isolate the effect of their different functional roles: the $\pi_{ta}$ agent is responsible for generating a candidate plan $\mathcal{P}$, whereas the $\pi_{ca}$ agent examines that plan according to the structured criteria defined by the $\mathcal{I}_{TCP}$. Thus, the distinction between the two agents is introduced through their prompts, responsibilities, and generation settings rather than through the use of different underlying LLM families.

\textbf{Generation settings:}~Because the two agents (\textit{$\pi_{ta}$ and $\pi_{ca}$ agents}) in the VST system perform different functions within the generation-critique-refinement cycle, we use different sampling temperatures for their LLM calls. Temperature controls the sharpness of the token sampling distribution: lower values produce more deterministic outputs, whereas higher values permit greater variation in generated responses. Following the task dependent effects of temperature reported by Renze et al.~\citep{renze2024effect}, we configured the $\pi_{ta}$ agent with a temperature of $T=0.3$. This relatively low but non-zero value was chosen to allow variation in candidate therapy plans while maintaining consistency with the therapeutic constraints specified in the $\mathcal{I}_\mathcal{TGP}$. In contrast, the $\pi_{ca}$ agent was configured with $T=0$, favoring deterministic and consistent responses when examining the candidate plan against the review criteria specified in the $\mathcal{I}_\mathcal{TCP}$~\citep{renze2024effect}. These settings reflect the complementary roles of the two agents: allowing controlled variation during candidate therapy plan generation while promoting consistency during the $\pi_{ca}$ agents’s self-review process.

The use of the same \textit{Gemini 3 Pro (Preview)} model family for both agents is important when interpreting the critique stage. The $\pi_{ca}$ agent should be understood as performing structured LLM self-review rather than providing an independent clinical or model based validation of the $\pi_{ta}$ agent's output. The critique therefore represents an additional computational review stage within the CITL framework, while final clinical judgment remains within the scope of SLP. In addition, no random seed was specified for the LLM calls; consequently, repeated executions may produce different candidate plans $\mathcal{P}$, particularly for the $\pi_{ta}$ agent given its non-zero temperature. For transparency, the prompts provided in \textit{Appendix~\ref{lab:appendix}} correspond to those used for the evaluation reported in this study.

}

\subsection{Implementation}
\label{sec:implementation}

The \textit{Virtual Speech Therapist} has already been developed as a web-based client--server application. The stuttering detection pipeline (\Cref{sec:sd}) is implemented in \textit{PyTorch}, with pre-trained model weights loaded via the \textit{Hugging~Face Transformers} library. Phonetic transcriptions are derived from speech segments using a wav2vec2-based phonemizer loaded through \textit{torchaudio}. Orthographic transcriptions are generated by \textit{Azure Speech Services} (Microsoft Cognitive Services), a cloud-based ASR module.

The multi-agent reasoning system described in \Cref{sec:agents} encompassing the $\pi_{ta}$ and $\pi_{ca}$ agents, and the iterative refinement loop is orchestrated with \textit{LangGraph}, a framework for building stateful, graph-based agent pipelines. \textit{LangChain} complements the stack with prompt management utilities and a unified interface through which both agents invoke the \textit{Gemini~3~Pro (Preview)} model (\Cref{sec:llms}).

The resulting user interface, data-flow architecture, and clinician approval workflow are detailed in \textit{Appendix} \ref{sec:ui_integration}.

\section{Discussion and Evaluation}

\textcolor{black}{
The qualitative clinical evaluation of VST generated therapy plans was conducted by a licensed SLP (G.L.), who reviewed ($\mathcal{D}_{SLP}$, Eq. \ref{eq:dec}) all generated therapy plans $\mathcal{P}$, with a phonetician (F.H.) providing input on phonetic aspects. Both are co-authors of this study and have clinical and research experience in fluency disorders. The evaluation examined the therapy plans $\mathcal{P}$ generated by the VST system; how the stuttering behaviours in each sample were interpreted, how therapeutic strategies were selected and sequenced, and how the VST system responded to clinician feedback ($\mathcal{R}$). Clinical outcomes, treatment efficacy and the impact of the system on everyday clinical practice were not assessed.} The recordings were used only to evaluate the VST system and were submitted to the system by the reviewing clinician; no client was present when the plans $\mathcal{P}$ were generated or reviewed ($\mathcal{D}_{SLP}$, Eq. \ref{eq:dec}), and no client received a plan. The evaluation therefore tested the VST system on existing recordings, not its use in a clinical encounter. \textcolor{black}{The 16 stuttered speech samples used for therapy plan generation were recorded in English and obtained from individuals who stuttered, aged 20–35 years, residing in Portugal. However, further demographic information, including sex and native language, was not available to us}. The therapy plans $\mathcal{P}$ were generated with the version of the system described here, in which the patient information entered in the interface is not passed to the language model; any age-related content in the generated plans was therefore inferred by the model and not derived from data. We cannot describe the sample by these characteristics, and we cannot assess whether the quality of the generated plans varies with speaker age.

\textbf{Evaluation Framework}. {\color{black}The clinician reviewed the generated therapy plans ($\mathcal{P}$) produced by the VST system after the $\pi_{ca}$ agent stage via \Cref{eq:dec,eq:mod}. Where changes were considered necessary, clinical feedback $\mathcal{R}$ was provided and the revised plan $\mathcal{P}$ was reviewed again. This allowed the review to examine how the VST agentic system incorporated corrections relating to the interpretation of stuttering behaviours, the selection of therapeutic strategies and their sequencing. Some of the therapy plans $\mathcal{P}$ reviewed are reproduced in \textit{Appendix} \ref{lab:appendixb}.
The review focused on whether the proposed strategies were appropriate for the stuttering behaviours identified in each recording, on the clinical rationale given for them, and on the sequencing of the intervention. Particular attention was given to the distinction between core stuttering behaviours, physical struggle and avoidance related behaviours, since these may call for different therapeutic responses.
The review was based on the speech recordings alone, without access to a comprehensive clinical assessment of the clients. No formal scoring system, independent comparator or inter-rater reliability assessment was used. The observations reported below therefore reflect the clinical judgment of one SLP and should be read within those limits.
}

{
\color{black}\textbf{Clinical Review of the Generated Therapy Plans}. The VST generated structured therapy plans ($\mathcal{P}$) that were generally understandable and drew on strategies commonly used in adult stuttering intervention, including \textit{stuttering modification} and \textit{fluency shaping}~\citep{guitar2018stuttering}, \textit{avoidance reduction therapy}~\citep{sisskin2022avoidance}, and approaches informed by CBT and ACT~\citep{menzies2008experimental, beilby2012acceptance}. The generated plans ($\mathcal{P}$) also provide explanations for the proposed strategies and their progression, making the rationale behind the suggested interventions explicit. Their appropriateness, however, could only be judged against the information available in the recordings, not against the clients' broader clinical needs. \textcolor{black}{This constraint was visible in at least one case, where the system’s output reflected the limits of what a recording can show. In that plan, the detection system $f_{SD}$ did not identify overt stuttering. Rather than focusing exclusively on fluency, the VST system indicated that a more detailed clinical assessment was needed to explore the possibility of covert stuttering, and proposed an intervention addressing cognitive and emotional aspects of the condition. Concealment is clinically relevant in its own right; higher self-rated covertness is associated with greater salience of stuttering and with lower quality of life, social role satisfaction and speech usage, independently of observable speech severity~\citep{boyle2025more}.}
\par

\par
This example illustrates the value of considering stuttering beyond its observable speech characteristics~\citep{yaruss2006overall, tichenor2019stuttering, jokar2026addressing}. The absence of overt stuttering in a recording does not, however, establish the presence of covert stuttering, and neither the speaker's emotional responses nor avoidance behaviours can be reliably determined from the audio sample alone. Both belong to the speaker's own account of the condition and require direct clinical assessment~\citep{tichenor2019stuttering}.
\par
In the judgment of the reviewing clinician, the activities proposed in the plans were generally clear, and several could be adapted to individual therapeutic goals and clinical contexts. Their acceptability to clients and their effect on clinical workload were not evaluated here.

}

{
\color{black}
\textbf{Limitations Observed in the Initial Therapy Plans}.
In four of the reviewed plans, fluency shaping techniques were initially introduced before addressing fear, avoidance, and physical struggle associated with stuttering. During the review, the SLP reordered these components, considering that desensitization and avoidance reduction should receive greater attention before introducing speech motor strategies in these cases. Here, desensitization refers to gradually and supportively addressing the fear and negative reactions associated with stuttering through exposure to stuttering related experiences, consistent with Van Riper's stuttering modification approach~\citep{vanriper1973treatment}. However, this observation should not be interpreted as supporting a fixed sequence of therapeutic interventions, as their selection and relative emphasis should reflect each individual's clinical presentation, goals, and preferences. This view is consistent with individualized, multidimensional approaches to stuttering therapy, which emphasize adapting interventions to the needs of the individual rather than following a single prescribed sequence~\citep{sonsterud2020works}.

A related limitation concerned the interpretation of speech behaviours by the stuttering detection model, $f_{SD}$. In one of the reviewed cases, interjections and pre-speech fillers were classified as prolongations, which influenced the direction of the proposed interventions. On reviewing the recordings, the clinician considered that some of these behaviours might instead represent avoidance responses associated with the anticipation or fear of stuttering, as described in studies of adults who stutter~\citep{tichenor2019stuttering, sisskin2022avoidance}. When such a distinction is overlooked, the resulting therapy plan may focus primarily on modifying speech production without adequately addressing the behaviours that may contribute to the individual's communication difficulties. Nevertheless, this alternative interpretation remains provisional, as it was based on the same recordings and would require confirmation through further assessment and discussion with the person who stutters.

The review also identified concerns regarding the number of exercises proposed during the initial phase of intervention. In two plans, more than five practice items were introduced in this phase, leading the clinician to consider whether presenting several new strategies simultaneously might place excessive demands on some clients. However, the actual demands experienced by clients were not assessed, and this observation therefore reflects the clinician's judgment rather than a measured effect. Furthermore, the prompts governing therapy plan generation do not currently specify a limit on the number of exercises that can be included in each phase, leaving the number and pacing of practice activities open to variation.

Taken together, these observations illustrate how limitations in automated speech classification and therapy plan generation can influence the interventions initially proposed. They also highlight the importance of clinician involvement in interpreting speech behaviours, considering alternative explanations, and selecting and adapting therapeutic strategies to the individual's needs. Although automated speech analysis can provide useful information to support therapy planning, its outputs must be interpreted in the broader context of a comprehensive assessment of the person who stutters and should not be treated as a substitute for clinical judgment.
}

{
{
\color{black}
\textbf{Clinician Feedback and Therapy Plan Revision}. In four of the reviewed plans, clinician feedback $\mathcal{R}$ prompted revisions to the interpretation of stuttering behaviors, the therapeutic focus, or the sequencing of interventions. In some cases, the initial plan $\mathcal{P}$ placed considerable emphasis on fluency shaping, whereas the recording suggested that fear, avoidance, or physical struggle might also warrant attention. In response to this feedback, the revised plans $\mathcal{P}$ incorporated additional strategies, including voluntary stuttering, desensitization, and stuttering modification. Voluntary stuttering, historically also referred to as pseudostuttering, involves the deliberate production of stuttering like disfluencies for therapeutic purposes~\citep{vanriper1973treatment}.

Beyond the selection of interventions, clinician feedback $\mathcal{R}$ also led to revisions in the proposed therapeutic goals. In some cases, the revised plans $\mathcal{P}$ shifted the primary focus away from fluency based measures toward reducing physical struggle, tension, and avoidance while promoting greater ease in communication. This shift reflects outcome frameworks that emphasize the speaker's experience of stuttering, rather than fluency alone, as an important dimension of therapeutic outcomes~\citep{yaruss2006overall, tichenor2020recovery}.
\par 
}
\textcolor{black}{The VST system incorporated the feedback $\mathcal{R}$ provided during SLP review ($\mathcal{D}_{SLP}$, \Cref{eq:dec,eq:mod}) and generated revised plans $\mathcal{P}$ that addressed the concerns raised. Although the revisions were considered responsive to the feedback $\mathcal{R}$ and the resulting plans $\mathcal{P}$ appeared to be of high quality, the initial and revised plans were not subjected to an independent, objective comparison under blinded conditions. Accordingly, the extent to which the revisions improved plan quality or the appropriateness of the proposed therapeutic recommendations remains undetermined. A controlled objective evaluation using independent assessment criteria is therefore warranted to establish the contribution of the multi-agentic refinement process and subsequent clinician feedback to the quality of the generated plans.}

}

\textbf{Representative Case Examples}.
{
\color{black}
Two of the cases summarized above are described in greater detail to illustrate how the interpretation of stuttering behaviours can influence the proposed therapeutic approach. In the first case, behaviours initially classified as prolongations were reconsidered during clinical review and interpreted as avoidance related interjections rather than core stuttering events. This distinction is important because the initial interpretation had led to a therapy plan $\mathcal{P}$ focused primarily on speech motor control. Following clinician feedback \(\mathcal{R}\), the revised plan \(\mathcal{P}\) placed greater emphasis on addressing fear and avoidance, incorporating strategies such as voluntary stuttering and desensitization.

The second case involved a recording containing syllable repetitions, for which the proposed speech motor strategies focused on supporting speech initiation and motor timing. Based on the available recording, these strategies appeared appropriate, and no further revision was requested. However, this agreement should not be interpreted as evidence that the assessment was comprehensive or that the proposed plan was suitable for the individual as a whole. Information about the speaker’s subjective experience of stuttering, emotional responses, communication participation, and personal goals was unavailable, yet such information would be necessary to assess the relevance of the proposed strategies to the person’s broader therapeutic needs.

Further examples of therapy plans $\mathcal{P}$ generated by the system are provided in \textit{Appendix} \ref{lab:appendixb}, illustrating the range of proposed strategies and their intended progression.
}

{\color{black}
\textbf{Limitations of the Evaluation and Future Directions}. Several limitations should be considered when interpreting the findings of this initial evaluation. First, the review was conducted by a single SLP with the input from a phonetician on phonetic aspects. Moreover, the generated plans $\mathcal{P}$ were not compared with independently developed clinician generated plans. The reviewer was aware that all plans were LLM generated, knew whether each plan $\mathcal{P}$ was an initial or revised version, and provided the feedback $\mathcal{R}$ used to produce the revisions. Consequently, the evaluation was not blinded and may have been influenced by reviewer bias, confirmation bias, and expectancy effects. Furthermore, the inherent non-determinism of LLMs may prevent the exact verbatim reproduction of generated plans $\mathcal{P}$, although subsequent generations are expected to remain broadly consistent with the original outputs. A second limitation concerns the information available to the VST system, as stuttering is a multidimensional condition whose impact extends beyond observable speech disfluencies~\citep{yaruss2006overall, iverach2014social}. In adults, this broader impact includes anticipation, avoidance, emotional responses, communication participation, and the effects of stuttering on daily life~\citep{tichenor2019stuttering}. Although these dimensions are essential to understanding an individual's experience of stuttering, they cannot be adequately captured by a short speech recording alone. As a result, their assessment requires direct clinical evaluation and input from the person who stutters. This limitation is also reflected in the stuttering detection component, $f_{SD}$, which relies exclusively on audio input and consequently cannot capture silent blocks or secondary behaviours, such as head movements. These behaviours are therefore not represented in the stuttering profile, $\mathcal{S}$, provided to the $\pi_{ta}$ agent, potentially limiting the contextual information available for therapy plan generation. Future work could address this limitation by incorporating multimodal based analysis or clinician recorded observations of these behaviours, thereby enabling a more comprehensive representation of the individual's stuttering profile.
Another limitation concerns the perspective represented in the therapy planning process. The current VST system is clinician directed and does not capture the person's own account of stuttering, including their attitudes, experiences, and the broader impact of stuttering on their life~\citep{constantino2022social, gerlachhouck2022interrupting}, as assessed by instruments such as the OASES~\citep{yaruss2006overall}. Future work should therefore involve adults who stutter in co-designing the VST system and ensure that therapy goals reflect their individual priorities and lived experiences, with greater emphasis on neurodiversity affirming and acceptance based approaches rather than treating fluency as the sole or primary goal~\citep{irani2025reimagining}.
\par

\textcolor{black}{Finally, preliminary observations suggest that incorporating critique feedback $\mathcal{R}$ from the $\pi_{ca}$ agent improves the quality of therapy plans $\mathcal{P}$ generated by the $\pi_{ta}$ agent. However, these observations remain qualitative and require a systematic objective evaluation by developing a new clinically-defined assessment rubric. Future studies should also investigate whether the choice of LLM family for the $\pi_{ca}$ agent influences the quality of the resulting plans $\mathcal{P}$. Such comparisons were beyond the computational scope of the present study, as evaluating multiple LLM families requires substantial computational resources. Ultimately, the VST framework must also be evaluated to determine whether its generated recommendations satisfy appropriate clinical safety requirements before they can be deployed in clinical practice.}

}

\section{Conclusion}

{
\color{black}
In this preliminary study, we introduced \textit{Virtual Speech Therapist}, a CITL, multi-agentic framework for developmental stuttering therapy planning.  The framework integrates a stuttering detection model $f_{SD}$ with a $\pi_{ta}$ agent that generates candidate therapy plans $\mathcal{P}$ and a $\pi_{ca}$ agent that iteratively reviews and refines them before an SLP makes the final decision  ($\mathcal{D}_{SLP}$, \Cref{eq:dec,eq:mod}) to approve, reject, or request revisions. A qualitative assessment of 16 recordings by one clinician provides preliminary support for the feasibility of VST system and its potential to assist therapy planning. Further studies are required to evaluate its clinical effectiveness, scalability, and potential impact on access to care, long-term outcomes, and quality of life.
 \par

Future studies will evaluate the framework in larger, more diverse cohorts with multiple independent clinicians, including comparative analyses and, where feasible, randomized controlled trials and ablation of the critique step. We would also focus on developing a clinically grounded rubric for evaluating therapy plans and extending the VST framework to other speech and language disorders.
}

\bmhead{Acknowledgements}

Declaration of generative AI and AI-assisted technologies in the manuscript preparation process. During the preparation of this work the author(s) used ChatGPT in order to finetune. After using this tool/service, the author(s) reviewed and edited the content as needed and take(s) full responsibility for the content of the published article.

\bmhead{Funding Declaration} This work did not receive any funding.

\section*{Data availability}
The SEP-28k-E dataset used in this study is available in a public repository at \url{https://github.com/th-nuernberg/ml-stuttering-events-dataset-extended}.
\bmhead{Ethics and Consent to Participate}
\textcolor{black}{The stuttering detection model was trained on the publicly available SEP-28k-E dataset. The 16 recordings of adults used to test the VST system had been made earlier in routine practice at the authors' clinic and were anonymized before use. The adults concerned had given their consent to the use of their recordings. No client took part in a session with the system or received a plan generated by it.}

\section*{Code availability}
The full Virtual Speech Therapist multi-agent codebase is available at \url{https://github.com/pmarmaroli/vocametrix-platform/tree/main/python/vstagent}. An interactive user interface for the proposed system is available at \url{https://vocametrix.com/ai/stuttering-therapy-planning-agent}, facilitating real time stuttering assessment and personalized therapy planning.

\newpage

\begin{appendices}

\section{User Interface and Clinical Integration}
\label{sec:ui_integration}
The VST system integrates a web-based frontend with a persistent \textit{Python} backend service to deliver therapy generation with clinician oversight. This section describes the user interface design, data flow, and approval workflow that bridges VST recommendations with clinical judgment.

\subsection{Frontend Design and Workflow}

The clinical interface supports two operational modes: a \emph{classification only} $f_{SD}$ mode for rapid stuttering analysis without therapy generation, and a \emph{full therapy generation} mode for comprehensive assessment and plan development.

The clinician begins by entering patient data including stuttered speech sample $\textbf{x}$ along with its metadata. In the version of the system used for the evaluation reported in \Cref{sec:vst_agent}, this information was stored with the session and included in the exported report, but only the language of the recording was passed to the language-model prompts; it was not included in the prompts of the $\pi_{ta}$ agent and the $\pi_{ca}$ agent. A later version of the code passes it to both agents, and its effect on the generated plans has not been evaluated.
\par
\textbf{Intended clinical context.} The clinician provides the recording (live or uploaded) and the clinical information, reviews the generated draft therapy plan $\mathcal{P}$, and decides (\Cref{eq:dec,eq:mod}) to approve, reject or modify it; a plan is released only after this decision. Whether the client is present during the recording or during the review of the plan is left to the clinician: the prototype does not prescribe it, and it was not studied here. The study did not measure the time or resources saved or added by the system. Generating a plan takes about 2 to 4 minutes of computation, to which the clinician's review time is added, and whether the total is shorter than writing a plan from scratch is unknown, so no saving can be claimed. Audio input may be either a live recording or an uploaded speech sample. Table~\ref{tab:segmentation_params} summarises the configurable audio segmentation parameters. This segmentation strategy ensures continuous temporal coverage of stuttering events by preventing artificial boundaries from disrupting event detection.

\begin{table}[ht]
  \centering
  \caption{Audio segmentation parameters available in the user interface.}
  \label{tab:segmentation_params}
  \begin{tabular}{lll}
    \toprule
    \textbf{Parameter} & \textbf{Options} & \textbf{Default} \\
    \midrule
    Segment duration & 3\,s, 4\,s, 5\,s & 4\,s \\
    Overlap ($k$) & 0\%, 25\%, 50\%, 75\% & 50\% \\
    \bottomrule
  \end{tabular}
\end{table}

\textbf{Data Flow and API Architecture:}~ The system implements asynchronous REST API endpoints to handle the computationally intensive therapy generation workflow (typically 120–240\,s). Table~\ref{tab:api_workflow} outlines the four-stage request lifecycle.

\begin{table*}
  \centering
  \caption{Asynchronous API workflow stages.}
  \label{tab:api_workflow}
  \begin{tabular}{lp{9cm}}
    \toprule
    \textbf{Stage} & \textbf{Description} \\
    \midrule
    Submission & Patient data and audio are submitted; the system returns a unique session identifier for asynchronous tracking. \\
    Processing & The backend analyses audio, performs chunked classification, and generates therapy recommendations. Real-time progress updates are streamed to the frontend. \\
    Results & A comprehensive therapy plan is displayed, including stuttering classification, therapeutic rationale, weekly progression, and intervention strategies. \\
    Export & HTML documentation is generated for clinical records and patient handouts. \\
    \bottomrule
  \end{tabular}
\end{table*}

\subsection{Results Presentation}

Once processing completes, the interface presents results through several expandable sections:

\begin{itemize}
  \item \textbf{Analysis Summary}: Audio characteristics (duration, sample rate, number of chunks) and an aggregate stuttering-type distribution visualised as a donut chart.

  \item \textbf{Overall Classification}: The primary and secondary stuttering types, weighted average confidence, a classified severity level (mild, moderate, severe), stuttering percentage across all chunks, and---when phonemic transcription is available---a list of phonemes correlated with disfluent regions (\emph{problematic phonemes}).

  \item \textbf{Chunk Heatmap}: A compact grid where each cell represents one audio chunk. Cell colour encodes the classified stuttering type and opacity reflects classification confidence. Hovering over a cell reveals the time span, type, confidence, phonemic transcription, and textual transcription; clicking a cell plays the corresponding audio segment directly in the browser.

  \item \textbf{Therapy Recommendations}: Primary goals, therapeutic rationale, week-by-week progression (typically 6--8 weeks), and specific evidence-based techniques with clinical reasoning traces linking observations to expected outcomes.

  \item \textbf{Generation History}: Records of all refinement iterations when modifications were requested during approval, including the prompts generated by the therapy and critic agents for full transparency.
\end{itemize}

\textbf{CITL Approval Workflow}:~ A structured three-action approval process, described in Section~\ref{sec:agents}, is required before a plan is marked as ready for delivery: the system does not release a plan without a clinician decision. Each decision, its timestamp and the clinician's feedback are recorded in the session and included in the exported report. This record is kept only for the duration of the session and in the exported document; the prototype has no long-term audit store.

\subsection{Backend Architecture}

The persistent backend service maintains machine-learning models and therapy agents in memory to eliminate model-loading overhead (30--120\,s per cold start). Table~\ref{tab:backend_components} describes its principal components.

\begin{table*}[ht]
  \centering
  \caption{Backend service components.}
  \label{tab:backend_components}
  \begin{tabular}{lp{9cm}}
    \toprule
    \textbf{Component} & \textbf{Role} \\
    \midrule
    Model Persistence & Speech-analysis models (wav2vec2 phonemiser, stuttering classifier) and VST agents are initialised at service startup and retained in memory across requests. \\
    Classification Caching & When a classification-only session transitions to full therapy generation, cached chunk classifications are reused. Only phonemic transcription and ASR are executed as a post-processing step, avoiding redundant ML inference. \\
    Asynchronous Coordination & Frontend and backend communicate through inter-process mechanisms with health monitoring to prevent cascading failures. \\
    \bottomrule
  \end{tabular}
\end{table*}

  \newpage
  \begin{userprompt}
  \section{Prompts}
  \label{lab:appendix}
    \subsection*{TherapyAgent Prompt}
    You are a \textit{LICENSED}, expert speech-language pathologist (SLP) specializing in stuttering disorders for: (\{language\_desc\}).

    You apply contemporary approaches:
    \begin{enumerate}
      \item Stuttering Modification (voluntary stuttering, pull-outs, cancellations).
      \item Fluency Shaping (easy onsets, light contacts, continuous phonation).
      \item Avoidance Reduction (ARTS) (desensitization, fear reduction).
      \item CBT/ACT/REBT (cognitive restructuring, acceptance).
    \end{enumerate}

    \textit{ROLE}: Clinical Decision Support. You do NOT provide medical diagnosis or replace human judgment.

    SAFETY \& ETHICS:
    \begin{itemize}
      \item \textit{STRICTLY AVOID}: Voice strain, breath holding, excessive practice, or shame-inducing language.
      \item \textit{PRIORITIZE}: Empowering, non-judgmental, person-centered language.
      \item  \textit{ADAPT}: Tailor vocabulary and examples to patient's age and cultural context.
    \end{itemize}

    \textit{RED FLAGS:
    }If patient indicates severe distress/trauma, include: "\textit{URGENT CLINICAL NOTE}: Patient profile indicates serious concerns requiring immediate human assessment. Recommend urgent consultation."

    OUTPUT REQUIREMENT:
    \begin{itemize}
      \item Output ONLY valid JSON following the schema below.
      \item NO markdown, NO explanatory text.
    \end{itemize}

    THERAPY PLAN JSON SCHEMA: \{schema\_str\}

    The schema includes a \texttt{clinicalReasoning} object inside each strategy, with the fields: \texttt{observation}, \texttt{clinicalRationale}, \texttt{expectedOutcome}, and \texttt{evidenceBase}.

    INSTRUCTIONS: \\
    Generate an evidence-based, personalized plan following ASHA guidelines.
    \begin{itemize}
      \item Target specific stuttering types with practical exercises.
      \item Progress from awareness to real-world application.
      \item Include measurable goals, home practice, and psychoeducation.
    \end{itemize}

    EXPLAINABILITY REQUIREMENT (CRITICAL): \\
    For EVERY strategy in the plan, you MUST provide a \texttt{clinicalReasoning} object that makes your decision-making transparent. This follows a Chain-of-Thought (CoT) Input--Thought--Output cycle:
    \begin{enumerate}
      \item \textbf{observation} (Input): What specific pattern in the patient's stuttering analysis motivates this strategy? \\
        \textit{Example: ``Patient exhibits 60\% block-type disfluencies concentrated on plosive consonants /p/, /b/, /k/ with visible jaw tension.''}
      \item \textbf{clinicalRationale} (Thought): Why is this strategy the clinically appropriate response? Reference the therapeutic approach. \\
        \textit{Example: ``Blocks on plosives indicate laryngeal tension. Light articulatory contacts (Fluency Shaping) reduce subglottic pressure and facilitate airflow, directly addressing the tension source.''}
      \item \textbf{expectedOutcome} (Output): What measurable improvement should result? \\
        \textit{Example: ``Reduction in block duration from $>$2\,s to $<$0.5\,s on target phonemes within 3 weeks, measured via timed readings.''}
      \item \textbf{evidenceBase}: Reference the clinical guideline or literature. \\
        \textit{Example: ``Guitar \& McCauley (2010), ASHA Practice Portal: Fluency Disorders''}
    \end{enumerate}
    This reasoning chain will be presented to the reviewing clinician for informed validation.

    Create a personalized therapy plan in JSON format for \\
    PATIENT INFO:
    \begin{itemize}
      \item Type: {classification.get('stutteringType', 'unknown')}
      \item Transcription: {classification.get('transcription', 'N/A')}
      \item Phonemes: {classification.get('phonemes', 'N/A')}
      \item Characteristics: {classification.get('characteristics', ['Not specified'])}
      \item Locale: {locale}, \#\textit{defines language of a speech sample}
      \item (Infer age/context from characteristics if not explicit)
    \end{itemize}

    \textit{Note: At runtime, the prompt is dynamically enriched with} \texttt{ACOUSTIC PROFILE} \textit{(chunk-level type distribution and stuttering percentage),} \texttt{PER-CHUNK ANALYSIS} \textit{(temporal breakdown with time ranges, types, confidence, transcriptions, and phonemes), and} \texttt{PHONEME-DISFLUENCY CORRELATION} \textit{(over-represented phonemes in disfluent segments) when these data are available from the classification pipeline.}

    STRATEGY:
    \begin{itemize}
      \item Analyze patterns (repetitions, blocks, tension) \& severity.
      \item  Select focus:
        \begin{itemize}
          \item Repetitions $\rightarrow$ Fluency Shaping.
          \item  Blocks/Tension $\rightarrow$ Stuttering Modification.
          \item  Mixed $\rightarrow$ Blended.
        \end{itemize}

      \item  Formulate Goals: Functional, linked to real contexts.
      \item  Design trajectory: Awareness $\rightarrow$ Isolated $\rightarrow$ Words/Phrases $\rightarrow$ Sentences $\rightarrow$ Real contexts.
      \item  Create Materials: Interest-based, phonetically aligned (e.g., vowels for easy onset).
      \item  Integrate avoidance reduction \& CBT (reframing beliefs).
      \item  Monitoring: Define tracking metrics (e.g., 0-9 ease scales).

    \end{itemize}

    REQUIRED CONTENT MAPPING:
    \begin{itemize}
      \item  `explanation.stutteringTypeDefinition`: Assessment of profile.
      \item  `explanation.therapeuticRationale`: Approach \& WHY. MUST end with combined warning: " IMPORTANT LIMITATION: This AI-generated plan is a decision support tool requiring review by a qualified SLP. It does not substitute for professional judgment." (Translate if locale is not en-US/en-GB, but ALWAYS include both cautions in one block).
      \item  `primaryGoal.rationale`: Psychoeducation (normalize stuttering, effectiveness > fluency).
      \item  `strategies.clinicalReasoning`: For each strategy, provide a complete CoT reasoning chain:
        \begin{itemize}
          \item observation: Link directly to patient's stuttering data (type, phonemes, confidence scores)
          \item clinicalRationale: Explain the clinical logic connecting observation to this specific exercise
          \item expectedOutcome: State a measurable, time-bound expected outcome
          \item evidenceBase: Cite the relevant clinical guideline or research
        \end{itemize}
      \item  `strategies.instructions`: For each strategy:
        \begin{itemize}
          \item Approach name (1-2 sentences)
          \item Purpose (2-3 sentences explaining clinical rationale)
          \item 5-8 Practice Items with detailed instructions:
            \begin{itemize}
              \item  Each item: 3-4 sentences covering context, execution, frequency, progression
              \item Include specific examples with sample words/phrases
              \item Specify duration, repetitions, and success criteria
            \end{itemize}
          \item Home Practice (detailed daily schedule with 0-9 rating scales)
          \item  Troubleshooting tips (2-3 common issues and solutions)
        \end{itemize}
\item Example: "STRATEGY: Easy Onset... PURPOSE: Reduce tension... PRACTICE: 1) ... 2) ... 3) ... HOME: Practice 5min, rate ease 0-9."

\end{itemize}

OUTPUT ONLY THE JSON OBJECT.
\subsection*{CriticAgent Prompt}

The \textit{CriticAgent} receives a two-part input: a \textit{system message} defining its role and a \textit{human message} containing the therapy plan and evaluation instructions.

\paragraph{System Message.}
You are a clinical quality assurance specialist for speech therapy (\{language\_desc\}).

Your role is to ensure therapy plans are:
\begin{itemize}
\item Clinically sound and evidence-based
\item Ethically appropriate and safe
\item Properly tailored to patient needs
\item Following professional association guidelines
\end{itemize}
Provide structured, constructive, and critical feedback focusing on safety, effectiveness, and evidence strength.

\paragraph{Human Message.}
THERAPY PLAN TO REVIEW: \{therapy\_plan\_json\}

TARGET CONDITION:
\begin{itemize}
\item Primary Stuttering Type: \{primary\_type\}
\item Chunk Distribution: \{type\_distribution\}
\item Stuttering Percentage: \{stuttering\_percentage\}\%
\item Phoneme-Disfluency Correlation: \{phoneme\_correlation\_summary\}
\end{itemize}

CRITICAL REVIEW INSTRUCTIONS: \\
Evaluate the therapy plan in detail, covering the following domains:

\begin{enumerate}
\item \textbf{Clinical Soundness} --- Assess whether the plan matches the patient's stuttering type and severity. Determine whether techniques are appropriate, realistic, and developmentally suitable. Highlight any misalignments or inappropriate strategies.

\item \textbf{Safety Concerns} --- Identify potential risks or contraindications (physical, vocal, or psychological). Evaluate whether the plan avoids strain, fatigue, or anxiety triggers.

\item \textbf{Evidence Strength} --- Check whether each technique is supported by peer-reviewed research or recognized guidelines (e.g., ASHA, IFA). Note the quality of evidence (strong, moderate, limited) for each intervention.

\item \textbf{Improvements Needed} --- Suggest specific, actionable changes to improve the plan. Include guidance on frequency, duration, intensity, or sequencing of exercises. Highlight areas for progress monitoring, home practice, and carryover strategies.

\item \textbf{Structure and Clarity} --- Evaluate whether the plan is well-organized, readable, and easy for clinicians or caregivers to follow. Suggest improvements for formatting, stepwise instructions, and clarity.

\item \textbf{Explainability and Reasoning Transparency} --- Verify that each strategy includes a complete \texttt{clinicalReasoning} object with \texttt{observation}, \texttt{clinicalRationale}, \texttt{expectedOutcome}, and \texttt{evidenceBase}. Assess whether the reasoning chain is logically sound: does the observation correctly link to the rationale? Are cited evidence bases legitimate and relevant? Flag any strategies where reasoning is missing, circular, or insufficiently grounded.
\end{enumerate}

OUTPUT FORMAT: \\
Provide your assessment in a structured, bullet-point format for each domain:
\begin{itemize}
\item Domain Name (e.g., Clinical Soundness):
  \begin{itemize}
    \item Observation: \ldots
    \item Strengths: \ldots
    \item Concerns: \ldots
    \item Recommendations: \ldots
  \end{itemize}
\end{itemize}
Repeat for all domains: Clinical Soundness, Safety Concerns, Evidence Strength, Improvements Needed, Structure and Clarity, Explainability and Reasoning Transparency.

\end{userprompt}

\begin{updatedPrompt}
\subsection*{Refinement Prompt}

When the \textit{CriticAgent} identifies areas for improvement, the therapy plan is refined using a dedicated refinement prompt. The \textit{TherapyAgent} receives the original plan together with the critic feedback and is instructed to address all flagged issues while preserving what is working well.

\paragraph{System Message.}
You are an expert clinical speech-language pathologist and therapist specializing in stuttering therapy (\{language\_desc\}).

Your task is to REFINE an existing therapy plan based on critic feedback.

REFINEMENT REQUIREMENTS:
\begin{enumerate}
\item Address ALL points raised in the critic feedback.
\item Maintain the therapeutic goals and overall approach.
\item Strengthen weak areas identified by the critic.
\item Add missing components or details.
\item Ensure all evidence-based recommendations are incorporated.
\item Keep what is working well, improve what needs enhancement.
\item Every strategy MUST include a \texttt{clinicalReasoning} object with: \texttt{observation}, \texttt{clinicalRationale}, \texttt{expectedOutcome}, and \texttt{evidenceBase}.
\end{enumerate}

\paragraph{Human Message.}
Refine the existing therapy plan to address the critic feedback below.

\begin{itemize}
\item PATIENT INFORMATION: \{patient\_info\}
\item ACOUSTIC PROFILE: \{acoustic\_profile\} \textit{(when available)}
\item PHONEME-DISFLUENCY CORRELATION: \{phoneme\_correlation\} \textit{(when available)}
\item EXISTING THERAPY PLAN: \{previous\_plan\_json\}
\item CRITIC FEEDBACK: \{critic\_feedback\}
\end{itemize}

Revise the therapy plan to directly address all points raised in the critic feedback. Preserve what is working well, and strengthen areas flagged for improvement. Maintain clinical soundness and safety throughout.

\subsection*{Human Revision Prompt}

When the clinician selects \textbf{Modification} during the approval phase (\Cref{sec:agents}), a dedicated human revision prompt integrates the clinician's feedback into the plan.

\paragraph{Prompt.}
You are refining an existing therapy plan based on clinician feedback.

IMPORTANT: Do NOT create a new plan from scratch. Instead, take the existing plan below and modify it according to the clinician's feedback.

\begin{itemize}
\item CLINICIAN FEEDBACK: \{human\_feedback\}
\item EXISTING THERAPY PLAN TO MODIFY: \{current\_plan\_json\}
\item TARGET CONDITION:
  \begin{itemize}
    \item Stuttering Type: \{primary\_type\}
    \item Phonetic textual content: \{phonemes\}
  \end{itemize}
\end{itemize}

Modify the existing plan to address the clinician's concerns while preserving the overall structure and elements that are working well.

\end{updatedPrompt}

\begin{aiprompt}

\section{Example Therapy Plans Generated by the Virtual Speech Therapist}
\label{lab:appendixb}
\subsection*{Therapy Plans Generated by the System (Examples, Not Delivered to Patients)}

\noindent\textbf{Note on terminology.} The plans below are the unedited output of the language model, and none of them was delivered to a patient. They use terms from the stuttering modification and fluency shaping traditions. \emph{Desensitization} means reducing fear and avoidance of stuttering by facing it in structured steps, for example by stuttering on purpose (\emph{voluntary stuttering}, also called \emph{pseudostuttering}). A \emph{preparatory set} is an anticipated moment of stuttering approached with a modification strategy, a \emph{pull-out} eases out of a stuttered word, and a \emph{cancellation} pauses after a stuttered word and repeats it easily~\citep{vanriper1973treatment, asha2026fluency}. \emph{Easy onset} and \emph{light articulatory contact} are gradual voicing and gentle articulatory postures used to reduce tension~\citep{asha2026fluency}.

\subsection*{Stuttering Type: Prolongation 1}
\par
\subsection{Explanation:}
\begin{itemize}
\item \textbf{Stuttering Type}: Prolongation involves the involuntary stretching of a sound (audible sound flow) while articulatory movement is temporarily suspended. It often indicates excessive supraglottic tension (lips/tongue) or laryngeal tension.

\item \textbf{Patient Characteristics}: The patient exhibits prolongations specifically on fricatives (e.g., \textipa{/S/} in 'she') and complex clusters. This indicates a need to prioritize Light Articulatory Contacts (LAC) over general Easy Onsets, as the tension is likely articulatory rather than purely laryngeal.

\item \textbf{Therapeutic Rationale}: The plan utilizes an Integrated Approach modified to prioritize desensitization and tension reduction. Stuttering Modification (Van Riper) is introduced first to address emotional reactivity and the fear-tension-struggle cycle. Once the patient can manage tension and reactivity, Fluency Shaping strategies (Webster/Guitar) are introduced to refine articulatory kinematics.
\end{itemize}

\subsection{Primary Goal}
\begin{itemize}
\item \textbf{Goal}: The patient will demonstrate increased ease of speech during stuttering moments by reducing physical tension and emotional reactivity, as measured by self-rating scales and clinician observation of struggle behaviors.
\item \textbf{Target}: Consistently low self-reported tension (e.g., < 3 on a 10-point scale) during disfluencies, rather than a specific fluency percentage.
\item \textbf{Baseline}: Patient currently exhibits uncontrolled prolongations on approximately 10-15\% of syllables with associated physical tension and no modification attempts.
\end{itemize}
\subsection{Step by Step Plan}
\begin{enumerate}
\item \textbf{Step 1: Foundation: Identification \& Proprioception}
  \begin{itemize}
    \item \textbf{Week Range:} Weeks 1-2
    \item \textbf{Objective:} Patient will identify physical tension sites (specifically articulatory vs. laryngeal) and differentiate between 'tense' and 'loose' speech production.
      \begin{itemize}

        \item \textbf{Proprioceptive Awareness \& Voluntary Prolongation}:
          Developing high-resolution awareness of the articulators to locate the specific point of tension during a prolongation.
  \item \textit{Instructions}: RATIONALE: You cannot relax a muscle if you don't know it's tense. We must map the mouth. DRILL ITEMS: 1) 'The Tension Scan': Prolong /s/ in 'Sun'. Scan your body. Is the tongue pressing the roof of the mouth hard? 2) 'Tense vs. Loose Contrast': Say 'She' with maximum pressure on the /sh/ (100\% tension). Then say it with lips barely touching (10\% tension). Repeat 10 times. 3) 'Voluntary Stretching': Intentionally stretch the first sound of 'She could buy' for 2 seconds without struggle. Keep the airflow moving. HOME PRACTICE: Read aloud for 5 mins/day using Voluntary Stretching. Rate anxiety (0-9). TROUBLESHOOTING: If you feel panic, shorten the stretch. Use a mirror to ensure lips aren't pressing hard.
\end{itemize}
\end{itemize}

\item \textbf{Step 2: Stuttering Modification \& Desensitization}
\begin{itemize}
\item \textbf{Week Range}: Weeks 3-5
\item \textbf{Objective:} Patient will reduce emotional reactivity and physical tension during blocks using Pull-outs and Preparatory Sets prior to attempting fluency shaping.
\begin{itemize}
  \item \textbf{Pull-outs \& Preparatory Sets (Van Riper)}
    Pull-out: Modifying tension *during* a block. Preparatory Set: Pre-modifying articulators *before* an anticipated block.
\item \textit{Instructions}: RATIONALE: We must address the struggle behavior before trying to shape fluency. These tools reduce the 'fight' response. DRILL ITEMS: 1) 'Pull-out': Fake a hard block on 'She'. Catch it. Reduce tension by 50\%. Slide out. 2) 'Preparatory Set': Anticipate a block on 'Minimum'. Pause. Relax the lips *before* starting. Say it with a 'Soft Start'. 3) 'Resilience': If you push through a block, stop. Say 'I pushed.' Re-do it with a Pull-out. HOME PRACTICE: Practice Pull-outs on the phrase 'She could buy a minimum'. TROUBLESHOOTING: Focus on the feeling of release, not just getting the word out.

\item \textbf{Desensitization \& Self-Disclosure:}
Reducing the emotional weight of stuttering to lower baseline tension.
\item \textit{Instructions}: RATIONALE: Lowering emotional reactivity makes physical modification easier. DRILL ITEMS: 1) 'Self-Disclosure': Practice saying 'I stutter, so I might take a minute' to a friend. 2) 'Voluntary Stuttering': Enter a low-stakes situation (e.g., asking a family member a question) and intentionally stutter on one word without tension. HOME PRACTICE: Log one instance of self-disclosure or voluntary stuttering daily.
\end{itemize}
\end{itemize}

\item \textbf{Step 3: Fluency Shaping \& Generalization}
\begin{itemize}
\item \textbf{Week Range:} Weeks 6-8
\item \textbf{Objective:} With tension reduced, patient will apply Light Articulatory Contacts (LAC) and Continuous Phonation to facilitate smoother forward flow.
\begin{itemize}
\item \textbf{Light Articulatory Contact (LAC)}:
Reducing the physical pressure of the articulators (lips/tongue) during consonant production to prevent the 'locking' of the prolongation.
\item \textit{Instructions}: RATIONALE: Now that the fear-tension cycle is managed, we refine the movement. Prolongations on /sh/ occur when the tongue presses too hard. DRILL ITEMS: 1) 'Feather /sh/': Say 'She'. Bring lips forward but imagine they are made of cotton. Barely make the sound. 2) 'Fricative Drills': Practice /f/, /s/, /sh/, /th/. Touch the articulators gently. 3) 'Negative Practice': Press hard on 'Shoe' (Old Way), then touch lightly on 'Shoe' (New Way). Feel the difference. HOME PRACTICE: 10 mins daily reading. Highlight all fricatives. Apply LAC to them.
\item \textbf{Continuous Phonation \& Real World Transfer}:
Connecting words smoothly to keep the voice box vibrating, applied in hierarchical situations.
\item \textit{Instructions}: RATIONALE: Keeping the voice 'on' prevents the system from resetting. We now apply this to real life. DRILL ITEMS: 1) 'Chaining': Say 'She-could-buy' as if it is one long word. 2) 'Hierarchy Level 1': Call a business recording. Use LAC and Chaining. 3) 'Hierarchy Level 2': Ask a stranger for the time using the target phrase 'She could buy...'. HOME PRACTICE: Complete one hierarchy task daily. MAINTENANCE: Create a plan to revert to Step 2 (Pull-outs) if tension spikes.
\end{itemize}
\end{itemize}

\end{enumerate}

\subsection*{Stuttering Type: Prolongation 2}
\par
\subsection{Explanation:}
\begin{itemize}
\item \textbf{Stuttering Type}: Prolongation involves the involuntary stretching of a sound (typically the initial phoneme) where airflow continues, but the articulators remain fixed in position for an extended period.

\item \textbf{Patient Characteristics}:  Analysis of 'It's fine, but sometimes' indicates a pattern of tension on fricatives (/s/, /f/). The patient maintains airflow but delays the vowel transition. This plan is framed as 'Phase 1: Acquisition \& Transfer' based on Barry Guitar's Integrated Approach.
\item \textbf{Therapeutic Rationale}:  Integrating Van Riper's Stuttering Modification (for tension) and Webster's Fluency Shaping (for airflow) addresses both the mechanics and the 'Iceberg' of stuttering (Sheehan). We utilize the OASES to track quality of life.  IMPORTANT: This plan focuses on acquisition; long-term stabilization may require follow-up.
\end{itemize}

\subsection{Primary Goal}
\begin{itemize}
\item \textbf{Goal}: The client will modify tension during prolongations using 'Light Contacts' or 'Pull-outs' in sentence-level speech, reducing the average duration of stuttering moments to <1 second.
\item \textbf{Target}: 80\% accuracy in self-correction AND Clinician-rated reduction of prolongation duration to <1 second.
\item \textbf{Baseline}: Client exhibits prolongations on 15\% of syllables (\%SS) in conversation, with an average duration of 2.0 seconds per moment of stuttering.
\end{itemize}
\subsection{Step by Step Plan}
\begin{enumerate}
\item \textbf{Step 1: Identification \& Desensitization (Safety First)}
\begin{itemize}
\item \textbf{Week Range}: Weeks 1-2
\item \textbf{Objective}: Increase proprioceptive awareness and reduce emotional reactivity to prolongations using a safety hierarchy.
\begin{itemize}
\item \textbf{Voluntary Prolongation with Video Feedback}:
Intentionally stretching sounds to neutralize fear (Desensitization) and calibrating self-perception using video.
\item \textit{Instructions}: APPROACH: Negative Practice / Desensitization (Byrd et al.). SAFETY PROTOCOL: Establish a 'Safe Word' (e.g., 'Pause'). If anxiety spikes >7/10, use the word to stop immediately and breathe. PRACTICE ITEMS: 1. Video Calibration: Record a 1-minute monologue. Watch it together. Identify 'real' tension vs. 'felt' tension. Often the stutter looks less severe than it feels. 2. The 'Slide': Intentionally stretch /s/ or /f/ for 3 seconds. Keep eye contact. Do this ONLY in the therapy room initially. 3. Contrast Drill: Produce 'Sun' with 100\% tension (hard), then 50\%, then 10\% (light). Feel the difference. HOME PRACTICE: Practice 'Sliding' in front of a mirror for 5 mins. Do NOT do this in public yet. Rate anxiety 0-9.
\end{itemize}

\end{itemize}
\item \textbf{Step 2: Fluency Shaping: Light Contacts}
\begin{itemize}
\item  \textbf{Week Range:} Weeks 3-5
\item \textbf{Objective}: Reduce articulatory pressure on fricatives to prevent the 'locking' mechanism (Webster/Perkins).
\begin{itemize}
\item \textbf{Light Articulatory Contacts}:
Touching articulators with minimum pressure to facilitate airflow.
\item \textit{Instructions}: APPROACH: Fluency Shaping. PURPOSE: Prevent the physical block before it starts. PRACTICE ITEMS: 1. The 'Feather' Concept: Touch the roof of the mouth for /s/ as lightly as a feather. 2. Word Drill: 'Five, Fine, Fish'. If the lip turns white, pressure is too high. Aim for a 'whisper touch'. 3. Carrier Phrases: 'I see a...' naming objects. Focus ONLY on the light contact sensation. HOME PRACTICE: Read a paragraph. Highlight /s/ and /f/ words. Apply light contacts. If speech sounds 'slurred', increase precision but NOT pressure.
\end{itemize}
\end{itemize}
\item \textbf{Step 3: Modification \& Bridging the Gap}
\begin{itemize}
\item Week Range: Weeks 6-7
\item \textbf{Objective}: Apply 'Pull-outs' in structured interactions before attempting real-world challenges.
\begin{itemize}
\item \textbf{The Pull-Out \& Structured Roleplay}:
Modifying tension during a block (Van Riper) within a controlled conversational script.
\item \textit{Instructions}: APPROACH: Stuttering Modification. PURPOSE: Bridge the gap between drills and spontaneous speech. PRACTICE ITEMS: 1. The Pull-Out: Fake a prolongation on 'Sssseven'. Catch it, reduce tension, and slide to the vowel. Do not stop and restart (Cancellation). 2. Scripted Roleplay: Clinician plays a 'Store Clerk'. Client asks specific questions from a script. Goal: Use 3 pull-outs intentionally. 3. Hierarchy Worksheet: Create a list of speaking situations ranked 1-10 (e.g., 1=Talking to pet, 5=Ordering coffee, 10=Public speech). HOME PRACTICE: Choose a level 2 or 3 situation from your Hierarchy (e.g., talking to a close friend). Aim for one intentional pull-out.

\end{itemize}
\end{itemize}

\item \textbf{Step 4: Generalization \& Maintenance}
\begin{itemize}
\item Week Range: Week 8
\item \textbf{Objective}: Transfer skills to high-load situations and establish a relapse management plan.
\begin{itemize}
\item \textbf{Real-World Transfer \& Relapse Planning}:
Testing skills in high-pressure contexts and planning for future fluency fluctuations.
\item \textit{Instructions}: APPROACH: Generalization. PURPOSE: To ensure skills survive outside the clinic and prepare for the chronic nature of stuttering. ACTIVITIES: 1. The Phone Call: Make a call to a recorded line (e.g., movie times) then a live person. Use a Light Contact on the first word. 2. Relapse Plan: Write down 'What do I do if I have a bad speech day?' (Answer: Go back to Step 1 - Voluntary Stuttering to reduce tension). 3. Support: Discuss joining the National Stuttering Association (NSA) for community support. HOME PRACTICE: Complete one 'Challenge Call'. Log the outcome not by fluency, but by successful management of tension.
\end{itemize}
\end{itemize}

\end{enumerate}

\subsection*{Stuttering Type: Silent Block 1}
\par
\subsection{Explanation:}
\begin{itemize}
\item \textbf{Stuttering Type}:  Re-evaluation of the audio sample ('We do not own a freshness') identifies the pauses not as linguistic rests, but as 'Silent Blocks.' This indicates a laryngeal or articulatory posture where airflow is completely stopped despite the attempt to speak.

\item \textbf{Patient Characteristics}: The patient exhibits 'Silent Blocking,' characterized by a stoppage of airflow and sound. This suggests high laryngeal tension (glottal closure) and likely significant anticipation anxiety. The silence represents the moment of highest physical struggle.
\item \textbf{Therapeutic Rationale}: Therapy must pivot from general 'fluency maintenance' to specific 'Airflow Management' and 'Tension Reduction.' The priority is to re-establish airflow during the silent moments to prevent the vocal folds from locking. We utilize the ICF Model to address the frustration associated with these silent struggles.
\end{itemize}

\subsection{Primary Goal}
\begin{itemize}
\item Goal: The patient will identify the onset of silent blocks and utilize airflow strategies to release laryngeal tension, converting 'struggled silence' into 'easy sound.'
\item Target: Reduction in duration of silent blocks; Self-rated physical tension during blocks < 3/10.
\item Baseline: Presence of silent blocks disguised as pauses; high laryngeal tension presumed.
\end{itemize}
\subsection{Step by Step Plan}
\begin{enumerate}
\item \textbf{Step 1: Airflow Initiation \& Laryngeal Release}
\begin{itemize}
\item Week Range: Weeks 1-2
\item \textbf{Objective}: To replace the 'holding breath' mechanism of the silent block with passive airflow and gentle phonation onset.
\begin{itemize}
\item T\textbf{he 'Pre-Voice' Exhale (Airflow Management):}
Teaching the patient to release a small amount of air *before* engaging the vocal folds to prevent the glottis from slamming shut (the silent block).
\item \textit{Instructions}: DOSAGE: 1x Weekly Clinical Session; 10 mins Daily Home Practice. [PATIENT INSTRUCTIONS] 1. Identification: Recognize the 'stuck' feeling in your throat during the silence. That is your vocal cords closing tight. 2. The Sigh: Practice a passive sigh. Feel the air leave your mouth without sound. 3. Application to 'We': The word 'We' (/w/) is a common blocking point. Sigh out air, then gently turn on the voice: 'hhh-We'. 4. Negative Practice: Intentionally hold your breath and try to say 'We'. Feel the block. Then, stop, release the air, and use the sigh method. [RATIONALE] Sound cannot travel without air. Silent blocks occur when you try to speak without airflow. We are prioritizing 'air over voice'. [TROUBLESHOOTING] If you feel dizzy, take a break. You are hyper-ventilating. Focus on *gentle* airflow, not forceful pushing.
\end{itemize}

\end{itemize}
\item \textbf{Step 2: Proprioception \& Soft Contacts for Blocks}
\begin{itemize}
\item Week Range: Weeks 3-5
\item \textbf{Objective}: To use light articulatory contacts to navigate through blocks without forcing, specifically targeting the transition out of silence.
\begin{itemize}
\item \textbf{Light Contacts \& Pull-outs}:
Using reduced pressure on consonants to prevent the build-up of intra-oral pressure that sustains the block.
\item \textit{Instructions}: DOSAGE: 1x Weekly Clinical Session; 15 mins Daily Home Practice. [PATIENT INSTRUCTIONS] 1. Target /f/ in 'Freshness': If you block on the /f/, do not push. Lightly touch your lip to your teeth and let air hiss through. 'ffff-reshness'. 2. Target /d/ in 'Do': If you block silently on 'Do', visualize your tongue tip barely touching the roof of your mouth. Whisper the 'd' first. 3. The 'Pull-out': When you feel a silent block, do not push through it. Stop. Release tension by 50\%. Re-attempt the word with a 'Light Contact'. [RATIONALE] Fighting a block increases tension and prolongs the silence. 'Softening' the contact breaks the neurological loop of the block. [TROUBLESHOOTING] Ensure you are not just whispering. You need to engage the voice, but only *after* the air is moving.

\end{itemize}
\end{itemize}

\item \textbf{Step 3: Desensitization to Silence}
\begin{itemize}
\item Week Range: Weeks 6-8
\item \textbf{Objective}: To reduce the panic associated with the silence of a block, allowing the patient to remain calm enough to apply strategies.
\begin{itemize}
\item \textbf{Voluntary Pausing \& Reality Testing:}
Intentionally inserting silence into speech to desensitize the fear response, proving that a pause is not a catastrophe.
\item \textit{Instructions}: DOSAGE: 1x Weekly Clinical Session; Real-world application tasks. [PATIENT INSTRUCTIONS] 1. Intentional Silence: In a safe conversation, pause for 2 seconds before saying a word. 'We do not... [2 sec pause]... own.' 2. Observation: During the silence, keep your throat loose. Do not lock up. Watch the listener. Did they interrupt? 3. Reframing: A silent block is involuntary, but a pause is voluntary. By practicing voluntary pauses, you gain control over the silence. 4. Video Feedback: Record yourself. Notice that the silence often feels longer to you than it sounds to the listener. [RATIONALE] Panic tightens the throat, worsening the block. If you are not afraid of the silence, your throat will remain open, reducing the frequency of blocks. [TROUBLESHOOTING] If the silence triggers a real block, use your 'Airflow' strategy (Step 1) to ease out of it.
\end{itemize}
\end{itemize}

\end{enumerate}

\subsection*{Stuttering Type: Silent Block 2}
\par
\subsection{Explanation:}
\begin{itemize}
\item \textbf{Stuttering Type}: Blocks involve a temporary stoppage of airflow and voicing due to laryngeal or articulatory tension. While the patient history notes prolongations, current observation indicates blocks are the primary feature, characterized by struggle and silence before sound initiation.

\item \textbf{Patient Characteristics}:  Clinician observation identifies blocks as the dominant behavior (e.g., on plosives in 'pay', 'talk', 'can'). The patient likely exhibits high physical tension/struggle during these moments. Desensitization is now prioritized to reduce the fear of 'getting stuck'.

\item \textbf{Therapeutic Rationale}:  Based on clinician feedback, the hierarchy is inverted to place Stuttering Modification *before* Fluency Shaping. Addressing the struggle behavior (blocks) and fear (desensitization) first is crucial to prevent the patient from forcing speech. The sequence is now: Desensitization/ID → Stuttering Modification (Unlearning struggle) → Fluency Shaping (Learning ease) → Stabilization.

\end{itemize}

\subsection{Primary Goal}
\begin{itemize}
\item Goal: The client will demonstrate the ability to reduce physical tension during blocks using Pull-outs and Cancellations in 80\% of perceived stuttering moments, following a reduction in avoidance behaviors.
\item Target: 80\% successful tension reduction; Reduced Avoidance Scale score
\item Baseline: Client exhibits hard blocks with struggle behaviors on approx. 15\% of syllables; high reactivity to stuck moments.
\end{itemize}
\subsection{Step by Step Plan}
\begin{enumerate}
\item \textbf{Step 1: Phase 1: Identification \& Desensitization}
\begin{itemize}
\item Week Range: Weeks 1-2
\item \textbf{Objective}: Identify the physical sensation of 'blocking' and reduce the panic response through voluntary stuttering.
\begin{itemize}
\item \textbf{Mapping the Block:}
Locating exactly where the airflow stops (Lips for 'Pay', Tongue for 'Talk', Larynx for 'I').
\item \textit{Instructions}: APPROACH: Identification. PURPOSE: To separate the 'block' from the 'struggle'. ACTIVITIES: 1. Freeze Technique: When you block on 'Talk', freeze instantly. Don't push. Ask: 'Is my tongue glued to the roof of my mouth?' 2. Tension Rating: Rate the block 0-10. If it's a 9, voluntarily release it to a 5 without trying to say the word yet. 3. The Mirror: Watch a block happen. Observe if eyes blink or neck tightens. Acknowledge it neutrally.
\item \textbf{Voluntary Stuttering (Desensitization)}:
Intentionally mimicking blocks to break the fear-tension cycle.
\item \textit{Instructions}: APPROACH: Avoidance Reduction. PURPOSE: To prove that blocking is not catastrophic. ACTIVITY: In a safe environment, read a sentence. Intentionally insert a light block on 'Pay' or 'Can'. Hold it for 2 seconds, then let go. Notice that you are in control of the duration.
\end{itemize}
\end{itemize}
\item Step 2: Phase 2: Stuttering Modification (Cancellations \& Pull-outs)
\begin{itemize}
\item Week Range: Weeks 3-6
\item \textbf{Objective}: Learn to modify the tension of the block *after* or *during* the moment, prioritizing struggle reduction over perfect fluency.
\begin{itemize}
\item \textbf{Cancellations (Post-Block Correction):}
Pausing after a hard block to re-attempt with loose articulators.
\item \textit{Instructions}: APPROACH: Stuttering Modification. PURPOSE: To erase the motor plan of 'pushing' and replace it with 'releasing'. STEPS: 1. Block on 'Pay'. Finish the word with struggle. 2. STOP. Pause for 3 seconds. Exhale. 3. Re-say 'Pay' with the lips barely touching (light contact). RULE: You are not allowed to repeat the word until you have fully exhaled and relaxed the tension site.
\item \textbf{Pull-Outs (In-Block Release)}:
Catching the block in real-time and voluntarily relaxing the tension to slide into the vowel.
\item \textit{Instructions}: APPROACH: Stuttering Modification. PURPOSE: To stop the struggle immediately. STEPS: 1. Feel the block on 'Talk'. 2. Do NOT push harder. Instead, stabilize the tongue posture. 3. Voluntarily reduce tension (melt the ice). 4. Slide slowly into the vowel \textipa{/A/}. METAPHOR: Easing off the gas pedal when the wheels spin in mud.
\end{itemize}
\end{itemize}
\item \textbf{Step 3: Phase 3: Fluency Shaping (Light Contacts \& Easy Onsets)}
\begin{itemize}
\item Week Range: Weeks 7-8
\item \textbf{Objective}: Introduce fluency shaping techniques now that fear and struggle have been managed.
\begin{itemize}
\item
\textbf{Light Articulatory Contacts (For Consonant Blocks)}
Touching articulators together gently to prevent the buildup of intra-oral pressure that causes blocks.
\item \textit{Instructions}: APPROACH: Fluency Shaping. PURPOSE: Prevention of blocks on plosives (/p/, /b/, /t/, /k/). TARGETS: 'Pay', 'Talk', 'Can', 'Maybe'. INSTRUCTION: Imagine your lips/tongue are sore. Touch them together as lightly as possible to produce the sound. Reduce the 'explosion' of air.
\item \textbf{    Easy Onsets (For Vowel Blocks)
}    Gradual initiation of voicing for vowel-initial words.
\item \textit{Instructions}: APPROACH: Fluency Shaping. PURPOSE: To prevent laryngeal locking on words like 'I', 'About'. INSTRUCTION: Exhale a tiny amount of air (silent /h/) before engaging the vocal folds. 'h-I', 'h-About'. Smooth out the start.

\end{itemize}
\end{itemize}

\item \textbf{Step 4: Phase 4: Stabilization \& Transfer}
\begin{itemize}
\item Week Range: Weeks 9-10
\item \textbf{Objective}: Generalize strategies to complex sentences and daily life.

\begin{itemize}
\item
\textbf{    Hierarchy of Transfer:
}    Moving from clinical practice to real-world application.
\item \textit{Instructions}: APPROACH: Generalization. STEPS: 1. Phone call to a recorded line (low pressure). 2. Asking a question to a stranger. 3. Presentation/Conversation. TASK: Use a Pull-out intentionally during a conversation. Report back on the feeling of control.
\item \textbf{    Maintenance Plan
}:    Preparing for relapse and high-stress days.
\item  \textit{Instructions}: APPROACH: Long-term Management. GUIDANCE: If blocks return with high tension, stop Fluency Shaping (Phase 3). Return immediately to Phase 1 (Desensitization) and Phase 2 (Cancellations). Do not fight the block; analyze it.

\item

\end{itemize}
\end{itemize}

\end{enumerate}


\subsection*{Stuttering Type: Repetition 1}
\par
\subsection{Explanation:}
\begin{itemize}
\item \textbf{Stuttering Type}:  The patient presents with specific 'initial syllable repetitions' at the start of utterances, despite a general classification of fluency. The clinician notes an absence of anxiety, ruling out 'Covert Stuttering' or Social Anxiety. This profile suggests a specific motoric difficulty with the initiation of phonation and articulation, rather than a psychological avoidance mechanism.

\item \textbf{Patient Characteristics}:  Unlike the previous assumption of high internal monitoring, the patient is not anxious but struggles with the motor execution of starting sentences. The behavior is characterized by repetitive attempts to initiate the first syllable. The focus shifts from desensitization (treating fear) to fluency shaping (treating the motor lag), specifically targeting the initiation of speech using the target phonemes \textipa{/U/} and \textipa{/S/}.
\item \textbf{Therapeutic Rationale}: Since anxiety is absent, Cognitive Restructuring and Voluntary Stuttering are removed as they are unnecessary. The plan pivots to 'Fluency Shaping' techniques. The primary objective is to smooth the transition from silence to speech (initiation) to prevent the initial syllable repetition. Techniques like Easy Onsets and Preparatory Sets are indicated.
\end{itemize}

\subsection{Primary Goal}
\begin{itemize}
\item Goal: The client will eliminate initial syllable repetitions and achieve smooth utterance initiation using fluency shaping techniques.
\item Target: Frequency of initial syllable repetitions (measured per 100 syllables) \& Client self-rating of initiation effort (0-10).
\item Baseline: Frequent repetition of the initial syllable at the beginning of utterances; low anxiety.
\end{itemize}
\subsection{Step by Step Plan}
\begin{enumerate}
\item \textbf{Step 1: Motor Initiation Strategies}
\begin{itemize}
\item Week Range: Weeks 1-2
\item \textbf{Objective}: Establish smooth initiation of phonation to replace syllable repetitions at the start of utterances.
\begin{itemize}
\item \textbf{Easy Onsets (Gentle Initiation)}:
Gradually increasing vocal fold tension to start a word, preventing the 'hard' attack or repetition loop often seen at the start of sentences.
\item \textit{Instructions:} PURPOSE: To smooth the transition from silence to speech, preventing the 'stuttering block' or repetition loop at the start. MECHANISM: Exhale slightly before speaking, then gently turn on the voice like a dimmer switch, not a light switch. PRACTICE ITEMS (Targeting \textipa{/S/} and \textipa{/U/}): 1. Inhale, exhale slightly, then gently slide into: {`Shhh...ould'} (\textipa{/SUd/}). 2. `Shhh...ugar' (\textipa{/SUg@r/}). 3. `L...ook' (\textipa{/lUk/}). 4. `P...ut' (\textipa{/pUt/}). ACTION: Practice 20 sentence starters. Focus ONLY on the first word. If a repetition occurs, stop, exhale, and try the Easy Onset again.
\item \textbf{Rhythmic Initiation:}
Using a rhythmic cadence to assist in the timing of utterance onset.
\item \textit{Instructions}: PURPOSE: To provide an external timing cue that overrides the internal motor lag causing the repetition. PRACTICE: Use a finger tap or a metronome beat to trigger the start of the sentence. EXERCISE: Tap your leg once, and coordinate the onset of your voice exactly with the tap. Do not speak *after* the tap; speak *on* the tap. TARGETS: 'Should we go?', 'Look at that.', 'Push it over.' (Focus on the \textipa{/U/} and \textipa{/S/} sounds in the initial position).
\end{itemize}
\end{itemize}
\item \textbf{Step 2: Stabilization \& Generalization}
\begin{itemize}
\item Week Range: Weeks 3-4
\item \textbf{Objective}: Integrate initiation techniques into natural speech and complex sentences.
\begin{itemize}
\item \textbf{Preparatory Sets:}
Mentally rehearsing the motor plan for the first word before attempting to speak, ensuring articulators are in place.
\item \textit{Instructions}: APPROACH: Before speaking, pause for 1 second. 1. SCAN: Identify the first sound of the sentence (e.g., \textipa{/S/} in 'Should'). 2. PLAN: Physically place your tongue/lips in the position for that sound without turning on your voice. 3. ACT: Initiate airflow gently (Easy Onset) through that pre-formed shape. GOAL: To replace the 'repetition' habit with a 'planning' habit.
\item \textbf{Continuous Phonation (Linking):}
Keeping the voice 'on' between words to maintain momentum after the successful initiation.
\item \textit{Instructions}: PURPOSE: Once the initial word is successfully started without repetition, maintain the flow to prevent downstream disfluencies. TECHNIQUE: Imagine your voice is a violin bow that doesn't leave the string. Connect the end of the first word to the start of the second. EXAMPLE: 'Shhhould\_we\_go.' (Instead of 'Should. We. Go.'). APPLICATION: Practice reading short paragraphs, focusing heavily on the Easy Onset of the first word and the Linking of the subsequent words.
\end{itemize}

\end{itemize}

\end{enumerate}
\subsection*{Stuttering Type: Repetition 2}
\par
\subsection{Explanation:}
\begin{itemize}
\item \textbf{Stuttering Type}:  Sound repetition involves the involuntary repetition of a specific phoneme (e.g., 's-s-s-supermarket'). In this profile, it manifests as multiple iterations of the initial sound, indicating a temporary disruption in the transition to the next sound.
\item \textbf{Patient Characteristics}: The patient exhibits repetitions on vowels ('I') and voiceless fricatives ('s'). This indicates laryngeal tension (hard starts) and articulatory tension (tongue pressure).
\item \textbf{Therapeutic Rationale}:  This plan follows an Integrated Approach (ASHA Practice Portal). Per clinician feedback, the sequence has been adjusted to prioritize Stuttering Modification (Pull-outs) to manage active tension first, followed by Fluency Shaping (Light Contacts, Easy Onsets) to prevent stutters. It aligns with Yaruss \& Quesal’s ICF Model by addressing both the impairment (fluency) and the reaction (cognitive/emotional). IMPORTANT: This is a decision support tool. Clinical judgment is required.
\end{itemize}

\subsection{Primary Goal}
\begin{itemize}
\item Goal: The client will manage sound repetitions using 'Pull-outs', 'Light Articulatory Contacts', or 'Easy Onsets' in variable contexts with a self-rated Severity Rating (SR) averaging < 3/9, and demonstrate reduced negative reaction to stuttering.
\item Target: Spontaneous speech accuracy ~70\%; Daily Severity Rating (SR) < 3/9; Reduced avoidance behaviors.
\item Baseline: Uncontrolled repetitions (3-4 iterations) on initial sounds; Presumed moderate tension; No current modification strategies.
\end{itemize}
\subsection{Step by Step Plan}
\begin{enumerate}
\item \textbf{Step 1: FFoundation: Awareness \& Desensitization}
\begin{itemize}
\item Week Range: Weeks 1-2
\item \textbf{Objective:} Client will identify physical tension sites and reduce fear of stuttering through desensitization.
\begin{itemize}
\item \textbf{Proprioceptive Awareness \& Voluntary Stuttering:}
Learning to feel articulatory tension and reducing the emotional charge of stuttering.
\item \textit{Instructions}: APPROACH: Sensory Awareness / Desensitization. PURPOSE: You cannot change what you do not feel. We must identify 'tight' vs. 'loose' muscles. PRACTICE ITEMS: 1. TENSION SCANS: Tense tongue against roof of mouth (Level 10), then release (Level 0). Notice the contrast. 2. THE 'BOUNCE' STUDY: Intentionally repeat 's-s-s' with high tension, then with loose muscles. 3. VOLUNTARY STUTTERING: Purposefully stutter on 'Supermarket' without struggle. SAFETY WARNING: Ensure you feel safe before attempting voluntary stuttering. If anxiety spikes > 7/10, stop and revert to private practice. Do not force this in high-stress situations yet. 4. SEVERITY RATING (SR): Introduce a 1-9 scale (1=no stuttering, 9=severe). Rate your speech daily to track patterns, not just 'good' or 'bad' days.
\end{itemize}
\end{itemize}
\item \textbf{Step 2: Stuttering Modification: Pull-outs}
\begin{itemize}
\item Week Range: Weeks 3-5
\item \textbf{Objective}: Client will learn to modify tension *during* a moment of stuttering (Pull-out).
\begin{itemize}
\item \textbf{Pull-outs (In-block Modification):}
Catching the stutter, stabilizing the tension, and sliding out of it voluntarily.
\item \textit{Instructions}: APPROACH: Stuttering Modification. PURPOSE: Before focusing on prevention, we must learn to handle the moment of stuttering. When you get stuck, don't push harder. Pull out. STEPS: 1. CATCH: Feel the 's' or 'I' repetition starting. 2. FREEZE: Hold the tension momentarily. Don't retreat, don't push. 3. RELEASE: Voluntarily reduce tension in the tongue or throat while keeping airflow moving. 4. SLIDE: Transition slowly to the next sound. PRACTICE: Fake a tense stutter on 'Supermarket', freeze on the 'S', relax the tongue, and say the word.
\end{itemize}
\end{itemize}
\item \textbf{Step 3: Fluency Shaping: Light Articulatory Contacts (LAC)}
\begin{itemize}
\item Week Range: Weeks 6-8
\item \textbf{Objective}: Client will apply LAC to voiceless fricatives (/s/) to prevent tension buildup.
\begin{itemize}
\item \textbf{Light Articulatory Contacts:}
Reducing contact pressure of the tongue to prevent 'bouncing'.
\
\item textit{Instructions}: APPROACH: Fluency Shaping. VISUAL ANALOGY: Imagine your tongue is a butterfly landing on a flower (the roof of your mouth). It should not stomp; it should land gently. PRACTICE ITEMS: 1. THE FEATHER TOUCH: Say /s/ with barely any pressure. Sustain 'sssss'. 2. SYLLABLE SLIDE: 'Saaa', 'Seee'. Smooth transitions. 3. WORD DRILL: 'Supermarket'. If you feel pressure > 2/10, stop and restart lighter. 4. CARRIER PHRASES: 'I see a...' focusing on the /s/. TROUBLESHOOTING: If speech sounds 'mushy', you are too light. Aim for distinct but gentle contact.
\end{itemize}
\end{itemize}

\item \textbf{Step 4: Fluency Shaping: Easy Onsets}
\begin{itemize}
\item Week Range: Weeks 9-10
\item \textbf{Objective}: Client will use Easy Onsets to manage vowel-initial words ('I') without hard glottal attacks.
\begin{itemize}
\item \textbf{Easy Onset:}
Gradual initiation of phonation to prevent laryngeal locking.
\item \textit{Instructions}: APPROACH: Fluency Shaping. VISUAL ANALOGY: A ramp vs. stairs. Slide your voice on (ramp) rather than jumping on (stairs/hard start). PRACTICE ITEMS: 1. THE 'H' TRICK: Place a tiny, silent 'h' before 'I'. 'h-I went'. 2. THE SLIDE: Inhale, and start speaking exactly as exhalation begins.  SAFETY WARNING: Do not use a 'whisper voice'. The goal is full resonance. If you feel lightheaded, you are exhaling too much air before speaking (hyperventilation risk). Shorten the 'h' significantly.
\end{itemize}
\end{itemize}

\item \textbf{Step 4: Generalization \& Maintenance}
\begin{itemize}
\item Week Range: Weeks 11-12
\item \textbf{Objective}: Transfer strategies to variable contexts and establish a long-term maintenance plan.
\begin{itemize}
\item \textbf{Hierarchy of Transfer \& Cognitive Resilience:}
Moving from clinical tasks to real-world resilience.
\item \textit{Instructions}: APPROACH: Generalization / CBT / Maintenance. HIERARCHY TASKS: 1. VARIABLE CONTEXTS: Practice strategies with background noise (TV on), while walking, or on a phone call. 2. COGNITIVE REFRAMING: Identify thoughts like 'I must be perfect.' Replace with 'I can communicate effectively even if I stutter.' 3. LISTENER STRATEGIES: Teach family members to maintain eye contact and wait (do not finish sentences). MAINTENANCE: - RELAPSE PLAN: Stuttering is cyclical. If tension returns, go back to Week 1 (Proprioception) for 2 days. - SUPPORT: Consider joining a peer group (e.g., National Stuttering Association) for long-term support.
\end{itemize}
\end{itemize}

\end{enumerate}



\subsection*{Stuttering Type: Interjection 1}
\par
\subsection{Explanation:}
\begin{itemize}
\item \textbf{Stuttering Type}:  A 'block' is a moment where airflow and voicing are temporarily stopped due to excessive muscle tension. In this context, interjections are identified as secondary avoidance behaviors used to postpone the feared block.

\item \textbf{Patient Characteristics}:  The patient exhibits blocks on initiation ('Sometimes I...') and utilizes interjections as avoidance strategies. This indicates that the core issue is anticipatory fear and threat perception rather than solely a motor breakdown.
\item \textbf{Therapeutic Rationale}:  Based on clinician feedback, the plan is restructured to prioritize cognitive-emotional processing and desensitization over immediate fluency shaping. Interjections are treated as avoidance markers. Motor techniques are introduced only after fear is reduced, with Stuttering Modification preceding Fluency Shaping.
\end{itemize}

\subsection{Primary Goal}
\begin{itemize}
\item Goal: The patient will reduce avoidance-driven interjections and manage moments of blocking with decreased fear, tension, and cognitive load, demonstrating increased tolerance and adaptive responses during speech initiation across structured and spontaneous communication contexts.
\item Target: 1. Reduction of avoidance-driven interjections by 50\% in structured tasks. 2. Ability to identify and differentiate tension levels in 80\% of blocks. 3. Successful use of Pull-outs (adaptive release) in 70\% of identified blocks.
\item Baseline: Patient currently uses interjections to avoid blocks on vowel initiations and experiences high fear/tension. Baseline severity via SSI-4; Cognitive impact via OASES.
\end{itemize}
\subsection{Step by Step Plan}
\begin{enumerate}
\item \textbf{Step 1: Cognitive-Emotional Foundation \& Desensitization}
\begin{itemize}
\item Week Range: Weeks 1-4
\item \textbf{Objective:} Increase awareness of interjections as avoidance behaviors and build tolerance for speech variability through voluntary stuttering.
\begin{itemize}
\item \textbf{    Psychoeducation \& Avoidance Recognition
}    Understanding the role of threat perception and identifying interjections as safety behaviors that maintain anxiety.
\item \textit{Instructions}: APPROACH: Cognitive Awareness. CONCEPT: Interjections (like 'um', 'uh' before 'Sometimes') are 'safety behaviors' that briefly lower anxiety but keep the fear alive. PRACTICE ITEMS: 1. The Avoidance Log: Track moments where you used a filler word to delay saying a feared word. Note the feeling immediately before (e.g., 'Panic'). 2. Neutral Observation: When you hear yourself say 'um... Sometimes', pause. Acknowledge it internally: 'That was an avoidance strategy.' Do not judge it, just label it. 3. The Pause: Instead of filling the silence with an interjection, practice simply pausing. Allow the silence to exist.
\item  \textbf{Voluntary Stuttering (Desensitization):}
Intentionally stuttering to reduce the fear of the block and the urge to avoid.
\textit{Instructions}: APPROACH: Desensitization/Approach Behavior. PURPOSE: To prove that stuttering is not a catastrophic event and to reduce the need for interjections. PRACTICE ITEMS: 1. Voluntary Repetition: On 'Sometimes', intentionally bounce on the 'S' (S-s-s-sometimes) with eye contact. Do not hide it. 2. The 'Clean' Block: Intentionally block on 'I' without using an interjection first. Hold the silence for 2 seconds, then say the word. 3. Reflection: After a voluntary stutter, ask: 'Did anything bad happen?' (Answer is usually no).
\end{itemize}

\item
\end{itemize}
\item \textbf{Step 2: Stuttering Identification \& Differentiation}
\begin{itemize}
\item Weeks 5-8
\item \textbf{Objective:  }Develop in-the-moment awareness of the speech mechanism and differentiate tension levels without urgency to fix them.
\begin{itemize}
\item
\textbf{Exploration of the Speech Mechanism:}
Systematic identification of laryngeal involvement and airflow during blocks.
\textit{Instructions}: APPROACH: Proprioception. FOCUS: Where is the tension? (Larynx/Throat vs. Lips). PRACTICE ITEMS: 1. Tension Mapping: Replicate the block on 'I'. Is the throat tight or loose? Rate the tension 1-10. 2. Differentiation: Produce a 'Hard Block' (10/10 tension) on 'Sometimes'. Then produce a 'Soft Block' (3/10 tension). Feel the difference in the larynx. 3. Duration Awareness: Identify if a block is 'Long' (>2s) or 'Short' (<1s). Just observe.
\item \textbf{    Stuttering Modification: The Pull-out:
}    Introducing agency within the block to reduce struggle, framed as an adaptive response rather than a correction.
\item  \textit{Instructions}: APPROACH: Stuttering Modification. GOAL: To increase flexibility, not to force fluency. PRACTICE ITEMS: 1. Catch the Block: When stuck on \textipa{/aI/} (I), catch yourself pushing. Stop pushing, but don't back up. 2. Release Tension: While holding the articulatory posture, consciously lower laryngeal tension by 50\%. 3. Slide Out: Slowly transition into the vowel. 'S... (release tension)... ometimes'. NOTE: If you use an interjection, stop. Reset. Try the Pull-out again.

\end{itemize}
\end{itemize}
\item \textbf{Step 3: Supportive Fluency Shaping \& Generalization}
\begin{itemize}
\item Week Range: Weeks 9-12
\item \textbf{Objective}: Introduce Easy Onsets as supportive tools (not primary targets) and generalize adaptive responses to spontaneous speech.
\begin{itemize}
\item  \textbf{Easy Onsets (Supportive Tool):}
Using gentle airflow to facilitate initiation, introduced only after fear/avoidance has decreased.
\item  \textit{Instructions}: APPROACH: Fluency Shaping (Secondary). USAGE: Use this as a preventative tool when you anticipate high laryngeal tension, but do not use it to 'hide' stuttering. PRACTICE ITEMS: 1. Gentle Airflow: On vowel initiations like 'I' or 'Apple', start with a microscopic exhale before engaging the vocal folds. 2. Phrase Practice: 'h(silent)... I think so'. 3. Integration: If you feel a block coming, you have a choice: Voluntary Stutter OR Easy Onset. Both are acceptable.
\item  \textbf{Resilience \& Real-World Application:}
Applying desensitization and modification in spontaneous contexts.
\item  \textit{Instructions}: APPROACH: Generalization. PRACTICE ITEMS: 1. The 'No Avoidance' Challenge: Enter a conversation with the specific goal of NOT using interjections, even if it means blocking silently. 2. Disclosure: Tell a listener, 'I'm working on my speech, so I might pause or repeat sounds.' 3. Spontaneous Speech: Answer open-ended questions (e.g., 'What did you do today?') focusing on 'Agency'—handling blocks with Pull-outs rather than avoiding them.

\end{itemize}
\end{itemize}

\end{enumerate}
\subsection*{Stuttering Type: Interjection 2}
\par
\subsection{Explanation:}
\begin{itemize}
\item \textbf{Stuttering Type}:  Interjections and pre-speech fillers are avoidance behaviors used to postpone the initiation of a feared word. Unlike motor prolongations, these are behavioral strategies to delay speech due to anticipatory anxiety.

\item \textbf{Patient Characteristics}:  Patient exhibits interjections and fillers (e.g., 'uh', 'um', or schwa insertions like '\textipa{/thEn@/}') primarily before initiating speech. These function as safety behaviors to avoid the feared initial sounds (/th/, /n/).
\item \textbf{Therapeutic Rationale}:  The therapy focus shifts from motor execution (managing tension) to anxiety reduction and avoidance reduction. The goal is to eliminate the 'starter' sounds and encourage direct initiation of the target word, even if disfluent. We must decouple the fear of the word from the need to use a filler.

\end{itemize}

\subsection{Primary Goal}
\begin{itemize}
\item Goal: The patient will initiate target phrases directly without the use of antecedent interjections or fillers in 8 out of 10 trials during structured tasks, tolerating potential disfluency on the target word.
\item Target: 8/10 direct initiations (zero fillers) in structured reading
\item Baseline: Patient currently uses frequent interjections/fillers before initial fricatives and nasals to delay speech onset.
\end{itemize}
\subsection{Step by Step Plan}
\begin{enumerate}
\item \textbf{Step 1: Identification \& Desensitization of Avoidance}
\begin{itemize}
\item Week Range: Weeks 1-2
\item \textbf{Objective}: Identify the urge to use fillers/interjections and decouple the fear response from the target word.
\item  \textbf{The 'Clean Start' \& Voluntary Stuttering:}
Identifying the 'urge' to use a filler and choosing to stutter openly on the target word instead.
\item \textit{Instructions}: APPROACH: Identification \& Negative Practice. SAFETY PROTOCOL: Monitor anxiety. If the urge to say 'um' or 'uh' arises, pause. Do not speak until the urge subsides. PRACTICE ITEMS: 1. Identification: Read a sentence. Raise a finger every time you feel the urge to use a filler before a word. 2. Voluntary Stuttering on TARGET: Instead of saying 'Uh... then', go directly to 'Then' and intentionally stutter on the 'Th' (e.g., 'Th-th-th-then'). 3. Rationale: Show the brain that the word 'Then' is safe and does not require a shield (the filler). SUCCESS CRITERIA: Patient can identify the urge to use a filler and choose a voluntary stutter on the target word instead (10 trials).
\end{itemize}
\item \textbf{Step 2: Modification Phase I: Cancellation of Avoidance}
\begin{itemize}
\item Week Range: Weeks 3-4
\item \textbf{Objective:} Learn to stop immediately after an avoidance behavior (filler) is used and restart without it.
\begin{itemize}
\item \textbf{The Avoidance Cancellation}:
Stopping immediately if a filler is used, pausing to reset, and attempting the target word directly.
\item \textit{Instructions}: APPROACH: Behavioral Modification. PURPOSE: To break the habit of using fillers as a springboard into speech. INSTRUCTIONS: 1. If you say 'Uh... Then', STOP immediately. 2. Exhale completely. Wait 3 seconds. 3. Attempt the word 'Then' again directly. Do not use a run-up. 4. If you stutter on 'Then', that is acceptable. The goal is NO FILLER. PRACTICE ITEMS: 1. Word List: 'Then', 'This', 'No'. Deliberately use a filler ('Um-No'), stop, pause, say 'No' directly. SUCCESS CRITERIA: Patient successfully cancels a filler and re-attempts with a clean start in 8/10 trials.
\end{itemize}
\end{itemize}

\item \textbf{Step 3: Modification Phase II: Preparatory Sets }
\begin{itemize}
\item Week Range: Weeks 5-7
\item \textbf{Objective:} Pre-speech planning to reduce anticipatory anxiety and initiate airflow directly.
\begin{itemize}
\item \textbf{Preparatory Sets (The Silent Pause)}:
Replacing the audible filler with a silent, deliberate pause to prepare articulators for direct onset.
\item \textit{Instructions}: APPROACH: Stuttering Modification (Preparatory Sets). PURPOSE: To replace the 'panic' filler with a 'planning' pause. INSTRUCTIONS: 1. Before the feared word (e.g., 'Then'), PAUSE intentionally. 2. Do not say 'um'. Instead, visualize the articulators moving for the /th/ sound. 3. Start the airflow GENTLY before the voice. /h/ -> /th/. 4. Initiate the word directly. PRACTICE ITEMS: 1. Phrase practice: 'And... (silent prep)... then'. 2. Sentence completion: 'I will go to the... (silent prep)... theater.' SUCCESS CRITERIA: Patient utilizes a silent preparatory set instead of an audible filler in 8/10 sentence completions.
\end{itemize}
\end{itemize}

\item \textbf{Step 4: Direct Initiation \& Transfer}
\begin{itemize}
\item Week Range: Weeks 8-9
\item \textbf{Objective:} Maintain direct speech initiation (no fillers) in connected speech and real-world scenarios.
\begin{itemize}
\item \textbf{Direct Entry \& Tolerating Silence}
Practicing the tolerance of silence (rather than filling it) during conversation.
\item \textit{Instructions}: APPROACH: Generalization. PURPOSE: To become comfortable with the silence that occurs while formulating thoughts, reducing the need for fillers. PRACTICE ITEMS: 1. Timed Response: Partner asks a question. Patient must wait 2 seconds (in silence) before answering directly. NO 'Well...', 'Um...', or 'You know...'. 2. Transfer Task: Order a coffee. Focus entirely on starting the order directly (e.g., 'Can I have...' vs 'Um, can I have...'). 3. Anxiety Check: If you block on the word, let it happen. Do not retreat to a filler. SUCCESS CRITERIA: Patient completes 3 transfer tasks with zero reported avoidance fillers.
\end{itemize}
\end{itemize}

\item \textbf{Step 5: Maintenance \& Avoidance Monitoring}
\begin{itemize}
\item Week Range: Week 10
\item \textbf{Objective:} Long-term monitoring of avoidance behaviors to prevent the return of fillers.
\begin{itemize}
\item \textbf{Avoidance Audit \& Support}:
Self-monitoring tools to track if fillers are creeping back into speech patterns.
\item \textit{Instructions}: APPROACH: Relapse Prevention. INSTRUCTIONS: 1. The 'Filler Audit': Record a 1-minute conversation once a week. Listen back ONLY for interjections/fillers. If you hear more than 3, return to Week 3 (Cancellations). 2. Cognitive Reframing: Remind yourself that 'Silence is better than avoidance.' 3. Support: Discuss the fear of silence with a support group or mentor. SUCCESS CRITERIA: Patient demonstrates ability to self-analyze recordings and identify avoidance behaviors.
\end{itemize}
\end{itemize}

\end{enumerate}

\end{aiprompt}

\end{appendices}

\newpage
\bibliography{sn-bibliography}

\end{document}